\documentclass[lettersize,journal]{IEEEtran}
\usepackage{amsmath,amsfonts}
\usepackage{array}
\usepackage[caption=false,font=normalsize,labelfont=sf,textfont=sf]{subfig}
\usepackage{textcomp}
\usepackage{stfloats}
\usepackage{url}
\usepackage{verbatim}
\usepackage{graphicx}
\usepackage{cite}
\usepackage{multirow}
\usepackage{tabularx,booktabs}
\newcolumntype{C}{>{\centering\arraybackslash}X} 
\usepackage[dvipsnames]{xcolor}
\usepackage{longtable}
\usepackage{makecell, cellspace, caption}
\usepackage{array}
\usepackage{subcaption}
\usepackage{cuted}
\usepackage{multirow}
\usepackage[utf8]{inputenc}
\usepackage{tabularx}
\usepackage{array}
\usepackage{amssymb}
\usepackage{makecell}
\usepackage[margin=2cm]{geometry}
\usepackage{adjustbox}
\usepackage{multirow}
\usepackage{tabulary,booktabs}
\usepackage{ragged2e}
\usepackage{amsmath}
\usepackage{booktabs}
\usepackage{array}
\newcolumntype{L}{>{}l<{}}
\newcolumntype{C}{>{}c<{}}
\newcolumntype{R}{>{}r<{}}
\newcolumntype{P}{>{}p{5.1em}<{}}
\newcolumntype{F}{>{}p{4.5em}<{}}
\newcolumntype{A}{>{}p{4.9em}<{}}
\newcolumntype{B}{>{}p{3.3em}<{}}
\newcolumntype{D}{>{}p{1.8em}<{}}
\newcolumntype{E}{>{}p{3.8em}<{}}
\newcolumntype{G}{>{}p{1em}<{}}
\definecolor{c1}{RGB}{0,0,0}
\definecolor{c2}{RGB}{0,0,255}
\definecolor{c3}{RGB}{255,0,0}

\usepackage{amssymb}
\usepackage{pifont}
\usepackage{graphicx}
\usepackage{hyperref}
\hypersetup{colorlinks,allcolors=black}
\usepackage{mathtools}
\usepackage{xurl}
\usepackage{lipsum} 
\usepackage{tikz} 
\usepackage[linesnumbered,lined]{algorithm2e}
\usepackage{flushend}
\usepackage{mathtools}
\makeatletter
\renewcommand{\@algocf@capt@plain}{above}
\RestyleAlgo{ruled}
\usepackage{setspace}
\SetKw{KwParam}{Params:}
\SetKw{KwEnsure}{Ensure:}
\SetKw{KwSet}{Set}
\SetKw{KwInit}{Initialize}
\SetKw{KwDel}{Delete}
\SetKw{KwEn}{Ensure:}
\SetKwComment{KwComment}{::}{}
\let\oldnl\nl
\newcommand{\nonl}{\renewcommand{\nl}{\let\nl\oldnl}}

\makeatother
\usepackage{textcomp}

\def\B{
	\begin{bmatrix}
		D_{1,..,P} & \mathbf{1} \\ 
		\mathbf{1}^T & 0
\end{bmatrix}}
\SetKwProg{Fn}{Function}{:}{end}
\SetKwFunction{FEstimate}{\textnormal{$\text{Estimate}_{\text{SNR}}$}}
\SetKwFunction{FProj}{\textnormal{Projection}}
\begin{document}

\title{Hyperspectral Unmixing with 3D Convolutional Sparse Coding and Projected Simplex Volume Maximization}
\author{Gargi Panda, Soumitra Kundu, Saumik Bhattacharya, Aurobinda Routray,~\IEEEmembership{Member,~IEEE}

\thanks{Gargi Panda and Aurobinda Routray are with the Department of Electrical Engineering, IIT Kharagpur, India (email: pandagargi@gmail.com; aroutray@ee.iitkgp.ac.in).}
\thanks{Soumitra Kundu is with the Rekhi Centre of Excellence for the Science of Happiness, IIT Kharagpur, India (e-mail: soumitra2012.kbc@gmail.com).}
\thanks{Saumik Bhattacharya is with the Department of Electronics and Electrical Communication Engineering, IIT Kharagpur, India
	(email: saumik@ece.iitkgp.ac.in).}}

\markboth{Submitting journal }%
{Shell \MakeLowercase{\textit{et al.}}: A Sample Article Using IEEEtran.cls for IEEE Journals}


\maketitle

\begin{abstract}
Hyperspectral unmixing (HSU) aims to separate each pixel into its constituent endmembers and estimate their corresponding abundance fractions. This work presents an algorithm-unrolling-based network for the HSU task, named the 3D Convolutional Sparse Coding Network (3D-CSCNet), built upon a 3D CSC model. Unlike existing unrolling-based networks, our 3D-CSCNet is designed within the powerful autoencoder (AE) framework. Specifically, to solve the 3D CSC problem, we propose a 3D CSC block (3D-CSCB) derived through deep algorithm unrolling. Given a hyperspectral image (HSI), 3D-CSCNet employs the 3D-CSCB to estimate the abundance matrix. The use of 3D CSC enables joint learning of spectral and spatial relationships in the 3D HSI data cube. The estimated abundance matrix is then passed to the AE decoder to reconstruct the HSI, and the decoder weights are extracted as the endmember matrix. Additionally, we propose a projected simplex volume maximization (PSVM) algorithm for endmember estimation, and the resulting endmembers are used to initialize the decoder weights of 3D-CSCNet. Extensive experiments on three real datasets and one simulated dataset with three different signal-to-noise ratio (SNR) levels demonstrate that our 3D-CSCNet outperforms state-of-the-art methods.
\end{abstract}

\begin{IEEEkeywords}
3D convolutional sparse coding, 3D-CSC block, algorithm unrolling, projected simplex volume maximization, hyperspectral unmixing.
\end{IEEEkeywords}
\section{Introduction}
\IEEEPARstart{H}{yperspectral} image (HSI) captures hundreds of continuous spectral bands for each pixel, enabling the simultaneous extraction of detailed spectral and spatial information. With this capability, HSI has become a powerful tool in a wide range of applications, including vegetation and crop monitoring, mineral and geological exploration, land-use analysis, food safety inspection, and medical diagnostics \cite{physics}. However, a fundamental challenge in hyperspectral images is that the spectrum of each pixel is often a mixture of several materials. This spectral mixture phenomenon occurs due to multiple factors: the limited spatial resolution of HSI sensors, the close proximity and interaction of materials within a scene, and multiple photon reflections from layered or complex surfaces \cite{mixture1,mixture2,low_rank}. As a result, the observed pixel spectrum rarely corresponds to a single pure material. Separating the spectra of individual pixels into a set of spectral signatures (called endmembers), and determining the abundance fraction of each endmember, is an essential task in quantitative hyperspectral subpixel analysis. This process, denoted as hyperspectral unmixing (HSU), is currently used in many applications.

Over the years, a wide range of algorithms has been developed for the HSU problem. Traditional approaches model the spectral mixing process using either linear or nonlinear formulations. Under ideal conditions, the mixture can be effectively described by the linear mixing model (LMM), which assumes that the observed pixel spectrum is a weighted combination of endmember signatures \cite{lmm1}. In practice, however, real-world environments introduce spectral variabilities (SVs)—caused by illumination changes, atmospheric effects, material heterogeneity, and sensor noise—which significantly degrade the accuracy of LMM-based unmixing.
To address SVs, the perturbed LMM (PLMM) \cite{lmm2} models variability as an additive perturbation to each endmember. Yet, a single perturbation term is often insufficient to capture the full range of spectral distortions encountered in real scenes. More advanced LMM-based models \cite{lmm3,lmm4,lmm5} have therefore been proposed to better compensate for SVs, but linear models still struggle to represent complex nonlinear interactions between materials. To overcome the inherent limitations of linear formulations, Somers \textit{et al.} \cite{bm1} introduced the bilinear model (BM), which accounts for two-material interactions occurring in layered or intimate mixing scenarios. Fan \textit{et al.} \cite{bm2} further generalized the BM to handle a broader set of mixing patterns. However, BM-based methods typically assume interactions between only two materials within each pixel, which restricts their applicability. To address this limitation, Halimi \textit{et al.} \cite{bm3} proposed the generalized bilinear model (GBM), and subsequent improvements \cite{bm4,bm5} enhanced its performance, particularly for multi-layered mixing cases. Although these model-based traditional methods provide strong interpretability and physical grounding, they often rely on strict assumptions about specific mixing patterns. As a result, their effectiveness is limited when faced with the diverse and complex mixing behaviors observed in real-world hyperspectral scenes. 

\begin{figure}[t!]
	\centering
	\includegraphics[width=0.48\textwidth]{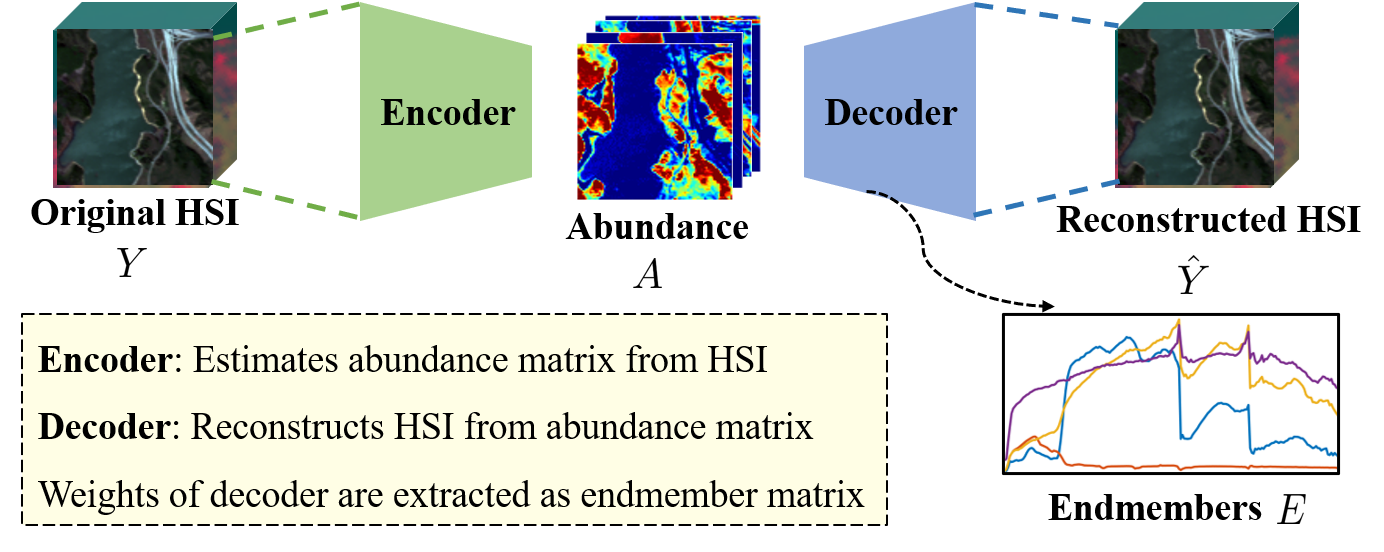}
	\caption{Illustration of AE framework used for the HSU task.}
	\label{fig:ae}
\end{figure}

In recent years, the rapid progress of deep learning (DL) has significantly advanced HSU performance. The high modeling capacity of deep neural networks (DNNs) enables them to capture complex nonlinear relationships, often achieving superior accuracy compared to traditional nonlinear models. Among these architectures, deep autoencoder (AE) models have emerged as the dominant framework for blind unmixing, primarily due to their unsupervised ability to learn compact, physically meaningful latent representations \cite{physics}. As illustrated in Figure \ref{fig:ae}, a typical AE consists of an encoder that estimates abundance fractions from the input HSI and a decoder that reconstructs the HSI, with the decoder weights extracted as the endmember matrix. Early AE methods \cite{ann1,AE1,AE2,AE3} utilized fully connected layers, which operate on a pixel-wise basis and therefore ignore the inherent 2D spatial correlations present in HSIs. To overcome this limitation, subsequent works \cite{cnn1,a2sn,dffn,bmae,acrnet,swcnet,ssafnet} introduced 2D convolutional layers to better exploit spatial contextual information. However, while 2D convolutions capture spatial patterns effectively, they may distort the intrinsic spatial–spectral coupling that characterizes hyperspectral data. To mitigate this issue, some studies \cite{3dcnn2,3dcnn} employed 3D convolutional layers, which more accurately preserve volumetric spatial–spectral structure. Although recent DL-based HSU methods demonstrate promising performance, their network architectures often have a ``black-box" nature, which makes it difficult to interpret the underlying abundance estimation process \cite{unrolling7_25}. 

To improve interpretability of the DL  methods, researchers \cite{unrolling5_20,unrolling6_22,unrolling3_22,unrolling1_23,unrolling2_23,unrolling4_23,unrolling7_25} have explored the deep unrolling networks. MNN-BU \cite{unrolling5_20} first unrolled the LMM model into a deep neural network using the iterative shrinkage thresholding algorithm (ISTA). Similarly, several other works \cite{unrolling6_22,unrolling3_22,unrolling1_23,unrolling2_23} have employed alternating direction method of multipliers (ADMM) algorithm to unroll the LMM. However, these methods are pixel-wise, and employ fully-connected layers to design the networks. To consider the 2D spatial relations in HSI, DIFCNN \cite{unrolling4_23} first employs 2D convolution layers in the unrolling-based networks. To further improve HSU performance, PnP-Net \cite{unrolling7_25} employs 2D dynamic convolution layers. However, such network designs ignore the 3D spatial-spectral information in the HSI. Also, the existing unrolling-based networks do not employ the AE framework, which is very effective for the unsupervised HSU task. Solely relying on the LMM can limit the modeling capacity of these unrolling networks.

\subsection{Motivation}
Despite the strong performance of recent DL-based HSU methods, they often have a ``black-box" nature, limiting their interpretability, particularly in relation to the underlying abundance estimation process. Unrolling-based DL methods improve interpretability by designing LMM-based network architectures. However, existing unrolling-based HSU networks do not leverage the enhanced modeling capability of the AE framework, limiting their ability to capture the complex nonlinear mixing effects present in real-world hyperspectral data. At the same time, effective HSU fundamentally requires the joint exploitation of spectral and spatial dependencies. The unrolling-based HSU methods typically treat these correlations separately or rely on 2D operations that overlook the intrinsic 3D structure of hyperspectral data. This highlights the need for an unrolling-based AE architecture equipped with 3D convolutional operations to better encode spectral–spatial interactions while maintaining interpretability through a principled network design. 

Another important factor influencing HSU performance is endmember initialization. The recent DL-based methods \cite{ssafnet,swcnet} mainly employ the vertex component analysis (VCA) method \cite{vca} to extract endmembers from the original HSI and use it for initialization. However, VCA exhibits suboptimal performance, particularly in the presence of noise \cite{noise_vca}. Moreover, VCA involves a random projection step that introduces variability in the estimated endmembers across different runs, and the
randomness of VCA is passed to the unmixing network \cite{a2sn,dffn}. These limitations indicate the need for an improved endmember initialization strategy.
\subsection{Contributions}
To achieve the above-mentioned goal, we propose an algorithm unrolling-based 3D convolutional sparse coding network (3D-CSCNet) for the HSU task. Unlike existing unrolling-based networks that do not exploit the enhanced modeling capacity of AE frameworks, our 3D-CSCNet is built upon an AE architecture. Our proposed 3D convolutional sparse coding (CSC) model represents the HSI using a 3D convolutional sparse representation, from which the abundance matrix is estimated. CSC, a widely used technique in image processing due to its interpretability and strong theoretical foundations, effectively extracts meaningful features under sparsity constraints \cite{fast_csc,lcsc,csc_2,fnet,unfolding}. Employing 3D CSC for HSI enables the joint learning of spectral and spatial relationships within the 3D HSI data cube.
Based on this model, we develop 3D-CSCNet, an autoencoder network designed for the HSU task. Since no prior work has introduced a learnable module to solve the 3D-CSC problem, we first propose a novel algorithm unrolling-based 3D-CSC block (3D-CSCB). In 3D-CSCNet, an initial abundance matrix is estimated using our 3D-CSCB, followed by a softmax operation. A pointwise convolution layer is then applied in the decoder to reconstruct the HSI, and the weights of this convolution layer are extracted as the endmember matrix.
Furthermore, we propose a projected simplex volume maximization (PSVM) algorithm for endmember estimation, and the estimated endmembers are used to initialize the decoder weights of 3D-CSCNet. Unlike VCA \cite{vca}, PSVM does not rely on random projections, ensuring no variation in the estimated endmembers across different runs. PSVM also provides improved endmember estimation performance, particularly in noisy conditions. Extensive experiments on three real datasets and one simulated dataset with three different signal-to-noise ratio (SNR) levels demonstrate that our proposed 3D-CSCNet has leading performance compared to the state-of-the-art (SOTA) methods.

Our contributions can be summarized as follows:
\begin{enumerate}
	\item We propose a novel 3D CSC model for abundance estimation. To the best of our knowledge, this is the first work to formulate abundance estimation using a 3D CSC model.
	\item To solve the 3D CSC problem, we design an algorithm unrolling-based 3D block (3D-CSCB). Building upon this, we introduce an end-to-end trainable AE network, named 3D-CSCNet, which employs our proposed 3D-CSCB to estimate the abundance matrix from the HSI input.
	\item We propose a novel projected simplex volume maximization (PSVM) algorithm for endmember estimation, and the estimated endmembers are used to initialize the decoder weights of 3D-CSCNet.
	\item We conduct extensive experiments on three real datasets and one simulated dataset across three different SNR levels. Our method has leading performance compared to the SOTA approaches.
\end{enumerate}

The remaining paper is organized as follows: Section \ref{method} presents the proposed method in detail. Section \ref{expeiments} describes the experimental setup, performance comparisons with te SOTA methods, and ablation studies. Finally, Section \ref{conclusion} concludes the paper.

\section{Methodology}
\label{method}

\subsection{Problem Formulation}
\label{sec_formulation}
The goal of HSU is to separate the endmembers and their corresponding abundances from a hyperspectral image (HSI). Given the HSI $Y\in \mathbb{R}^{L\times H\times W}$, where $L$ denotes the number of spectral bands, $H$ denotes the height, and $W$ denotes the width; we start with the widely used linear mixing model (LMM) to describe each pixel's spectrum \cite{swcnet}:
\begin{equation}\label{lmm_problem}
	Y=EA+\eta
\end{equation}

\noindent where $E\in \mathbb{R}^{ L\times P}$ represents the endmember matrix, $A\in \mathbb{R}^{ P\times H\times W}$ represents the abundance matrix of $Y$, $\eta\in \mathbb{R}^{L\times H\times W}$ represents the noise matrix, and $P$ denotes the number of endmembers. In practice, the abundance matrix must satisfy two physical constraints: the abundance non-negativity constraint (ANC), which requires all abundances to be non-negative, and the abundance sum-to-one constraint (ASC), which enforces that the abundances for each pixel sum to unity \cite{hsu_review}. 

Our approach is to design an interpretable and effective pipeline that extends beyond the linear assumptions of the LMM. We first estimate a sparse feature representation from the HSI, followed by the estimation of a preliminary abundance matrix using a learnable dictionary. A softmax mapping is then applied to obtain the final abundance estimates. The CSC-based encoder can incorporate nonlinear transformations that improve the non-linear modeling capability while remaining interpretable. The sparsity regularization selects only the salient features from the HSI, while the softmax function ensures that the ANC and ASC constraints are satisfied in a smooth and differentiable manner. Finally, following the auto-encoder framework \cite{AE1,AE2,AE3,swcnet},we apply a learnable linear function to the abundance matrix to reconstruct an HSI that closely matches the original, and extract the coefficients of this linear function as the endmember matrix.

To estimate the sparse feature representation from the HSI, we propose a 3D convolutional sparse coding (3D-CSC) model. Conventional 2D convolutional operations often distort spectral information, as they primarily capture spatial correlations while neglecting spectral continuity \cite{3dcnn2,3dcnn}. In contrast, 3D convolution can jointly learn spatial and spectral features, thereby preserving the intrinsic structure of hyperspectral data. We first reshape $Y$ to $Y\in \mathbb{R}^{1\times L\times H\times W}$, and formulate our 3D-CSC model as:
\begin{equation} \label{problem_3dcsc}
	Y = D(z)
\end{equation}

\noindent where $z\in \mathbb{R}^{C\times P\times H\times W}$ is the convolutional sparse feature estimated from $Y$, $C$ denotes the number of feature channels, and $D(\cdot)$ denotes a learnable 3D convolutional dictionary. Given the HSI $Y$, we first estimate the sparse feature $z$. We then estimate the preliminary abundance matrix $\hat{A}\in \mathbb{R}^{1\times P\times H\times W}$ using a 3D convolution operation, reshape it to $\hat{A}\in \mathbb{R}^{P\times H\times W}$, and apply a softmax operation to obtain the abundance matrix $A\in \mathbb{R}^{P\times H\times W}$. This process can be represented as:
\begin{equation} \label{eq_abundance}
	\begin{split}
	&\hat{A} = G(z)\\
		&A = \sigma(\hat{A}) 
	\end{split}
\end{equation}

\noindent where $G(\cdot)$ is a 3D convolutional  operation that generates the preliminary abundance estimate $\hat{A}$, and $\sigma(\cdot)$ is the softmax function enforcing the ANC and ASC constraints.

 In our formulation, we constrain the sparse feature representation $z$ using $\ell _1$-regularization. The feature $z$ is estimated by solving the following optimization problem,
\begin{equation}\label{eq_3dcsc}
	\underset{z}{\mathrm{Argmin}}\:\frac{1}{2}\Big|\Big|\:Y-D(z)\Big|\Big|_2^2 +\lambda ||z||_1 
\end{equation}
To solve Equation \ref{eq_3dcsc}, we propose a 3D-CSC block (3D-CSCB), which is described in the next subsection.

\begin{figure*}[hbt!]
	\centering
	\includegraphics[width=0.7\textwidth]{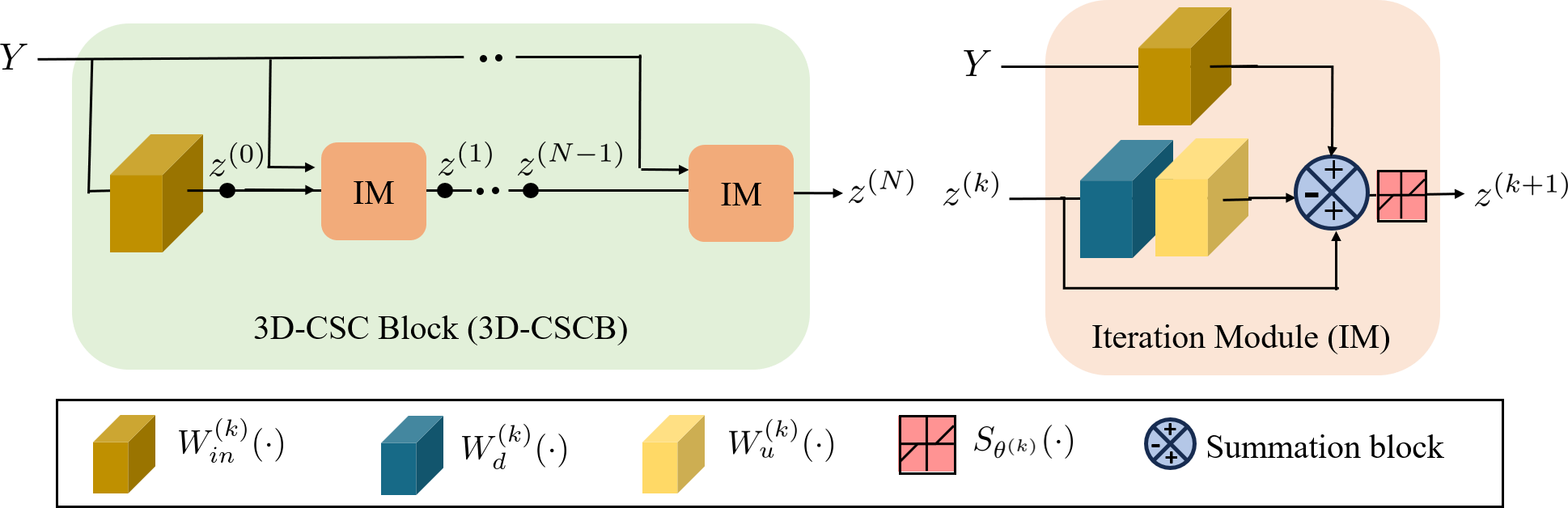}
	\caption{Structure of 3D convolutional sparse coding block (3D-CSCB).}
	\label{fig:cscb}
\end{figure*}
\subsection{Solving the 3D-CSC Problem}
We consider the 3D-CSC problem described in Equation \ref{eq_3dcsc}, where the feature $z\in \mathbb{R}^{C\times P\times H\times W}$ is estimated from the input $Y\in \mathbb{R}^{1\times L\times H\times W}$. 
Sreter \textit{et al.} \cite{lcsc} proposed a 2D convolutional extension of the learned iterative shrinkage thresholding algorithm (LISTA) \cite{lista} to solve the $\ell_1$-regularized 2D-CSC problem. In \cite{sinet}, the LCSC block \cite{lcsc} was improved for better sparse feature estimation. Motivated by these insights, we propose a 3D convolutional extension of \cite{sinet} for solving Equation \ref{eq_3dcsc}. Our iteration step for updating $z$ is given by,

\begin{equation}\label{eq_3dlista}
	\begin{split}
	z^{(k+1)}=S_{\theta^{(k)}}\Big(z^{(k)}&- W_{u}^{(k)}\Big(W_{d}^{(k)}\Big(z^{(k)}\Big)\Big)\\	
	&+W_{in}^{(k)}\Big(Y\Big)\Big)
	\end{split}
\end{equation}

\noindent
where $z^{(k)}$ is the estimation of $z$ at $k^{th}$ iteration step, $W_{u}^{(k)}(\cdot)$, $W_{d}^{(k)}(\cdot)$, and $W_{in}^{(k)}(\cdot)$ are learnable 3D convolution layers. Following \cite{sinet}, we use different operators and parameters at each iteration step to improve estimation accuracy. The soft thresholding function $S_{\theta^{(k)}}(\cdot)$ is defined as,

\begin{equation}\label{soft}
S_{\theta^{(k)}}(x)=sgn(x)\:max(|x|-\theta^{(k)}, 0)    
\end{equation}

\noindent where $sgn(\cdot)$ is the sign function and $\theta^{(k)}$ is a learnable threshold. As shown in Eqn. (\ref{soft}), $S_{\theta^{(k)}}(\cdot)$ nullifies input values whose absolute magnitudes are below $\theta^{(k)}$, thereby promoting sparsity and highlighting the salient features in the HSI. Since the HSI $Y$ has depth $L$, while  the sparse feature $z$ has depth $P$ with $P<L$, we use a strided convolution layer for $W_{in}^{(k)}(\cdot)$ in Eqn. (\ref{eq_3dlista}). The stride size along the spectral dimension is given by $ss=\left\lceil\frac{L}{P}\right\rceil$, while the stride size is unity along the spatial dimensions. Here, $\left\lceil\cdot\right\rceil$ denotes the ceiling function. We also constrain the threshold values $\theta^{(k)}$ to be non-negative and monotonically decreasing with the iteration index $k$, since the noise level in $z$ decreases over iterations. Inspired by \cite{sinet}, we constrain $\theta^{(k)}$ as,

\begin{equation} \label{eq_constrain}
	\begin{split}
		&\theta^{(k)}=sp(w_{\theta} k+b_{\theta})\:\:,\:\: w_{\theta} <0
	\end{split}
\end{equation}

\noindent
where $sp(\cdot)$ is the softplus function, $w_{\theta}$ and $b_{\theta}$ are learnable parameters. We unfold Eqn. (\ref{eq_3dlista}) into the iteration module (IM) shown in Figure \ref{fig:cscb}. Assuming $z^{(k)}=0; \:\: \forall k<0$, we stack multiple iteration modules to construct our 3D-CSC block (3D-CSCB).
\begin{figure}[t!]
	\centering
	\includegraphics[width=0.42\textwidth]{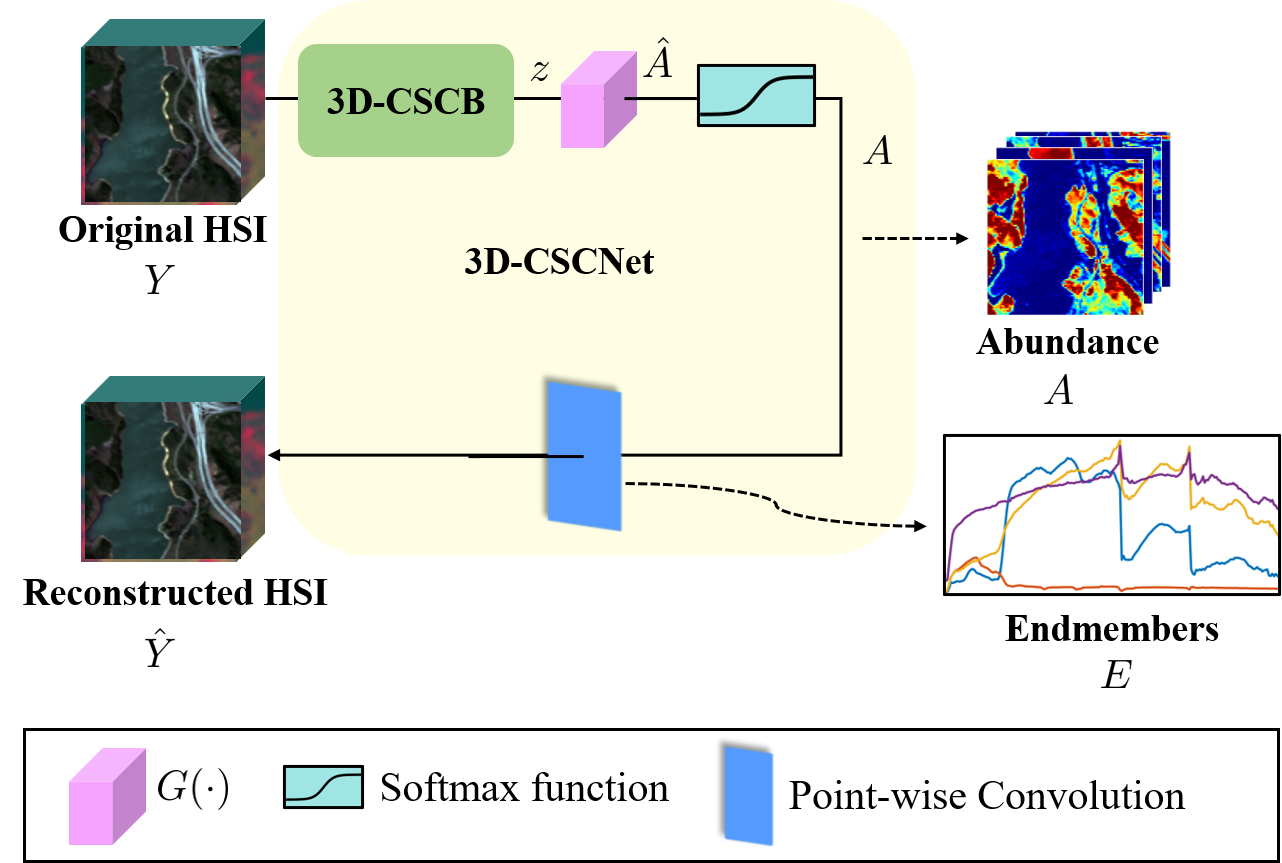}
	\caption{Overall architecture of 3D-CSCNet.}
	\label{fig:archi}
\end{figure}
\subsection{Network Architecture}
We propose an end-to-end trainable network, named 3D-CSCNet, for the HSU task. The 3D-CSCNet architecture is designed following the auto-encoder framework \cite{AE1,AE2,AE3,swcnet}, where the encoder estimates the abundance matrix and the decoder reconstructs an HSI—close to the original—using a linear layer. The weights of this linear layer are extracted as the endmember matrix. Figure \ref{fig:archi} illustrates the overall architecture of the proposed 3D‑CSCNet. Given the input HSI $Y$, the network first extracts a sparse feature representation $z$ through the proposed 3D-CSCB. This sparse feature is then passed through a learnable 3D convolutional layer $G(\cdot)$ to generate the the preliminary abundance $\hat{A}$. To ensure physical validity, a softmax function is applied to $\hat{A}$, yielding the final abundance map $A$, which satisfies the ANC and ASC constraints. Subsequently, a point‑wise convolution layer is applied to $A$ to reconstruct the HSI, denoted as $\hat{Y}$. The weights of this pointwise convolution layer are then extracted as the endmember matrix, enabling simultaneous estimation of both abundances and endmembers within a unified framework.  

Endmember initialization also plays a crucial role in HSU performance. The SOTA methods \cite{ssafnet,swcnet} employ the VCA method \cite{vca} to extract endmembers from the original HSI for initialization. However, VCA often performs sub-optimally in noisy conditions \cite{noise_vca} and includes a random projection step, which introduces variability in the estimated endmembers across different runs, and the
randomness of VCA is passed to the unmixing network \cite{a2sn,dffn}. To enhance performance and eliminate this variability, we propose a projected simplex volume maximization (PSVM) algorithm for endmember initialization. The PSVM algorithm is described in the next subsection.
\subsection{Projected Simplex Volume Maximization Algorithm}
\label{fsvm}
\begin{algorithm}[t!]
\setstretch{1.1}
\SetAlgoLined
\KwIn{$Y\in \mathbb{R}^{L\times H\times W},\:\:P$}

\KwOut{$E_i\in \mathbb{R}^{L\times P}$}
$N \gets H\cdot W$

$R \in \mathbb{R}^{L\times N} \gets \text{reshape}(Y)$

$\text{SNR}\gets \FEstimate(R,N,P)$

$\text{SNR}_{\text{th}}\gets 22+10\:log_{10}(P)$ 

\KwComment{Denoising for noisy HSI}

\If{\textnormal{SNR $<\text{SNR}_{\text{th}}$}}{
$J \gets \text{GaussFilt3D}(Y)$

$R \in \mathbb{R}^{L\times N} \gets \text{reshape}(Y)$
} 

$\text{SNR}\gets\FEstimate(R,N,P)$

\KwComment{Dimensionality Reduction}

\eIf{\textnormal{SNR $>\text{SNR}_{\text{th}}$}}{
$d \gets P$

$U_d,X_d \gets \FProj(R,N,d)$

$u \gets \text{mean}(X_d)$

$R_d \gets \frac{X_d}{\sum X_du}$

} 
{
$d \gets P-1$

$U_d,X_d \gets \FProj(R-\text{mean}(R),N,d)$


$R_d \gets  X_d$

}

\KwComment{find pixel indices using SVM}

indices $\gets \text{SVM}(R_d,P)\:$ 

\KwComment{Projecting back to $L$-dim. space}

\eIf{\textnormal{SNR $>\text{SNR}_{\text{th}}$}}{
$E \gets U_dX_d[:,\text{indices}]$
}
{
$E_i \gets U_dX_d[:,\text{indices}]+\text{mean}(R)$
}
\Fn{ $\textnormal{SNR} \gets $ \FEstimate{$R,N,P$}}{
$\bar{R} \gets \text{mean}(R)$

$U_P,R_P \gets \FProj(R-\bar{R},N,P)$

$P_R \gets \frac{\sum R^2}{N}$

$P_{R_P} \gets \frac{\sum R_P^2}{N}+\bar{R}^2$

SNR $ \gets$ $\Big|log_{10}\Big(\frac{P_{R_P}-\frac{P}{L}P_R}{P_R-P_{R_P}}\Big)\Big|$
}
\Fn{$(U_d, R_d)\gets$ \FProj{$R,N,d$} }{
$U_d \gets \text{SVDs}\Big(\frac{RR^T}{N},d\Big)$ 

$R_d \gets U_d^TR$ \
}
\caption{Projected Simplex Volume Maximization (PSVM)}\label{algo_psvm}
\end{algorithm}

We propose the PSVM algorithm to estimate the initial endmembers $E_i\in \mathbb{R}^{L\times P}$ from the HSI $Y$. Following \cite{vca}, our algorithm is designed under the LMM assumption, expressed as:

\begin{equation}\label{eq_unmixing}
	R = E_i \gamma \boldsymbol{\alpha}+\boldsymbol{n}
\end{equation}

\noindent where $R\in \mathbb{R}^{L\times N}$ is the reshaped HSI with $N = H\cdot W$ pixels, $\boldsymbol{\alpha}\in \mathbb{R}^{P\times N}$ is the corresponding abundance matrix, $\gamma\in \mathbb{R}$ is a tomographic modulation factor, and $\boldsymbol{n}\in \mathbb{R}^{L\times N}$ represents additive noise. Equation \ref{eq_unmixing} is similar to Equation \ref{lmm_problem}, except that the 3D HSI, abundance, and noise matrices are reformulated as 2D matrices. Additionally, the tomographic modulation factor $\gamma$ is considered but assumed to be unity, as in \cite{vca}.
The abundance matrix must satisfy the ANC and ASC constraints discussed in Section \ref{sec_formulation}. Under the idealized assumption of zero noise, the pixels of an HSI lie within a simplex whose vertices correspond to the endmembers. Consequently, the spectral vectors in $R$ span a subspace of dimension $P$, and the true endmembers lie at the vertices of the simplex with the maximum volume \cite{simplex1}. However, in practical scenarios, noise makes direct maximum-volume selection unreliable, often resulting in inaccurate endmember estimates. The proposed PSVM algorithm is specifically designed to address this issue. Algorithm \ref{algo_psvm} outlines the steps of the proposed PSVM method. Given $Y$, we first reshape it to $R$ in Step $2$. To identify the pixel indices that maximize the simplex volume, four operations are performed, as described below.

\subsubsection{Denoising}
  We perform denoising to suppress noise in the hyperspectral data. First, we estimate the SNR of $R$ using the function $\text{Estimate}_{\text{SNR}}(\cdot)$, defined in Steps $27$ to $33$. This function assumes that the meaningful signal lies in the $(L-P)$-dimensional subspace, while the noise lies in the $P$-dimensional orthogonal subspace. If the SNR of $R$ is above a threshold value $\text{SNR}_{\text{th}}$, no denoising is performed. Otherwise, we apply 3D Gaussian filtering to the original HSI volume $Y$ to obtain $J$, given by
   \begin{equation}
  	J = G*Y 
  \end{equation}
  where $*$ denotes convolution, and $G$ is the 3D Gaussian kernel defined as:
  \begin{equation}
  	G(x, y, z) = 
  	\frac{1}{(2\pi)^{3/2} \sigma_x \sigma_y \sigma_z} 
  	e^{-\frac{1}{2}\!\left(\frac{x^2}{\sigma_x^2} + \frac{y^2}{\sigma_y^2} + \frac{z^2}{\sigma_z^2}\right)}
  \end{equation}
  where \( \sigma_x, \sigma_y, \sigma_z \) denote the standard deviations of the Gaussian filter along each spatial dimension. The Gaussian filter smooths the HSI data, thereby removing high-frequency noise. The filtered output $J$ is then reshaped to match the shape of $R$. We follow \cite{vca}, which sets $\text{SNR}_{\text{th}}= 15+10\:log_{10}(P)$. However, our experiments show that a higher threshold value—defined in Step $4$ of Algorithm \ref{algo_psvm}—results in improved performance.
\subsubsection{Dimensionality Reduction}
After conditional denoising, we project $R$ onto a $d$-dimensional subspace, which represents the signal subspace under the LMM assumption. This projection preserves the essential simplex geometry \cite{vca}. Steps $10$ to $20$ in Algorithm \ref{algo_psvm} correspond to this dimensionality reduction, with the procedure differing depending on the estimated SNR. 

If the SNR exceeds $\text{SNR}_{\text{th}}$, we project $R$ onto the $P$-dimensional subspace using the $\text{Projection}(\cdot)$ function (Steps from $34$ to $37$). This function computes the eigenvectors $U_d$ of the correlation matrix corresponding to the largest $P$ eigen values, and calculates $X_d\in \mathbb{R}^{P\times N}$ as the projection of $R$ using $U_d$. Steps $13$ and $14$ then remove the tomographic modulation factor $\gamma$ in Equation \ref{eq_unmixing}, yielding the final projection $R_d\in \mathbb{R}^{P\times N}$.

When the SNR is below $\text{SNR}_{\text{th}}$, following VCA, we project $R-\bar{R}$ onto a $(P-1)$-dimensional subspace to compute $X_d$, where $\bar{R}$ is the mean of $R$. This improves the quality of the projected signal. We consider $X_d$ as the $d$-dimensional projection $R_d\in \mathbb{R_d}^{P-1\times N}$.

\subsubsection{Simplex Volume Maximization}
The $N$ pixels in $R_d\in \mathbb{R}^{d\times N}$, obtained after dimensionality reduction, lie within a simplex \cite{vca}. The $P$ endmembers correspond to the pixel indices that maximize the simplex volume. The simplex volume can be computed via the Cayley–Menger determinant. If the endmembers are $(e_1,...,e_P)$, and $d_{ij}$ is the squared Euclidean distance between $e_i$ and $e_j$, the simplex volume $V$ is given by \cite{simplex2},

\begin{equation}
\begin{split}
(-1)^P2^{P-1}((P-1)!)^2V^2&=det(C_{1,..,P})\\
C_{1,2,..,P} &= \B
\end{split}
\end{equation}

\noindent where, $D_{1,..,P}=[d_{ij}]_{i,j=1,..,P}$ is the squared distance matrix, $\mathbf{1}\in \mathbb{R}^P$ is the all-ones vector, and $det(\cdot)$ is the determinant operator. The simplex volume is iteratively maximized by replacing endmembers with pixels that increase the volume, repeating this procedure until no larger volume is found. The pixel indices corresponding to the final endmembers are recorded in Step $21$ of Algorithm \ref{algo_psvm}, and denoted as $\text{SVM}(\cdot)$.

\subsubsection{Projecting Back to Original Space}
The endmembers obtained by simplex volume maximization lie in the $d$-dimensional subspace. Finally, these pixels are projected back to the original $L$-dimensional space. Steps $22$ to $26$ of Algorithm \ref{algo_psvm} outline the process of projecting back the endmembers to the original space. 

Unlike VCA, the proposed PSVM algorithm does not involve any random projection. As a result, it ensures consistent endmember estimates across different runs with no variability.

\subsection{Two Stage Training Process}
In the HSU task, no ground-truth (GT) abundances or endmembers are available. In this unsupervised setting, a two-stage training strategy can improve performance \cite{ssafnet}.
We employ our 3D-CSCNet to reconstruct the HSI $\hat{Y}$ such that it closely approximates the original HSI $Y$. The training objective constrains $\hat{Y}$ to remain similar to $Y$ using the spectral angular distance (SAD) loss. To calculate the SAD loss, we first reshape $Y$ and $\hat{Y}$ to $R\in \mathbb{R}^{L\times N}$ and $\hat{R}\in \mathbb{R}^{L\times N}$, respectively, where $L$ is the number of spectral bands, and $N$ is the number of pixels. Then for the $j^{th}$ pixel, the SAD between reconstructed HSI $\hat{r}_j\in \mathbb{R}^{L\times 1}$ and GT $r_j\in \mathbb{R}^{L\times 1}$ is calculated as,

\begin{equation}\label{eq:loss_spectral}
	\text{SAD}(r_j,\hat{r}_j) = \text{cos}^{-1}\Bigg(\frac{<r_j, \hat{r}_j>}{||r_j||_2||\hat{r}_j||_2}\Bigg)
\end{equation}

\noindent where, $<\cdot, \cdot>$ and $||\cdot||_2$ denote the inner product and $L_2$ norm, respectively. Next we calculate the SAD loss for training as,

\begin{equation}\label{eq:loss}
	\mathcal{L} = \frac{1}{N}\sum_{j-1}^{N}\text{SAD}(r_j,\hat{r}_j)
\end{equation}

\noindent Training proceeds until convergence, after which the encoder output and the decoder convolutional layer weights are extracted as the estimated abundances and endmembers, respectively.

We initialize the decoder convolutional layer weights using our PSVM algorithm. Since PSVM provides accurate endmember estimates, we avoid training the decoder parameters from scratch. Instead, we adopt a two-stage training process. In Stage I, we train only the encoder parameters while keeping the well-initialized decoder layer fixed, ensuring that only the abundance matrix is updated with fixed endmembers. In Stage II, we train both the encoder and decoder parameters, enabling joint fine-tuning of abundances and endmembers.

\section{Experiments}
\label{expeiments}

To evaluate the performance of our proposed 3D-CSCNet, we have conducted extensive experiments. Section \ref{exp_datasets} describes the datasets in details. Then, we present the experimental setup, including evaluation metrics, training settings, and implementation details in Section \ref{exp_setup}. In Section \ref{exp_sota}, we report the quantitative and qualitative comparison of our proposed 3D-CSCNet with the SOTA methods. Finally, in in Section \ref{exp_ablation}, an ablation study is conducted to validate the effectiveness of our proposed method.  

\begin{table*}[t!]
\fontsize{7.4}{8}\selectfont
\centering
\caption{HSU performance comparison with the SOTA methods on three real data (Houston, Moffett, Jasper-Ridge) and one simulated data with three different SNR levels (10 dB, 20 dB, 30 dB). We highlight the best and second-best performances in  \textcolor{red}{\textbf{red}} and \textcolor{blue}{\textbf{blue}} colors, respectively.}
\begin{tabular}{D|B|P|FFFFFFA|E}
\hline
\multicolumn{3}{r|}{\multirow{2}{*}{Methods}} & \multicolumn{1}{c}{USTNet} & \multicolumn{1}{c}{A2SN} & \multicolumn{1}{c}{DFFN} & \multicolumn{1}{c}{HUMSCAN} & \multicolumn{1}{c}{BMAE} & \multicolumn{1}{c}{SSAFNet} & \multicolumn{1}{c|}{SWCNet} & \multicolumn{1}{c}{\multirow{4}{*}{\makecell{3D-CSCNet \\ (Ours)}}} \\
\multicolumn{3}{r|}{\multirow{1}{*}{}}& \multicolumn{1}{c}{\cite{ustnet}} & \multicolumn{1}{c}{\cite{a2sn}} & \multicolumn{1}{c}{\cite{dffn}} & \multicolumn{1}{c}{\cite{humscan}} & \multicolumn{1}{c}{\cite{bmae}} & \multicolumn{1}{c}{\cite{ssafnet}} & \multicolumn{1}{c|}{\cite{swcnet}} &  \\
\cmidrule{1-10}
\multicolumn{3}{r|}{Publications} & \multicolumn{1}{c}{TGRS} & \multicolumn{1}{c}{TGRS} & \multicolumn{1}{c}{TGRS} & \multicolumn{1}{c}{JSTARS} & \multicolumn{1}{c}{TGRS} & \multicolumn{1}{c}{TGRS} & \multicolumn{1}{c|}{TGRS} &  \\
\multicolumn{3}{r|}{Years} & \multicolumn{1}{c}{2023} & \multicolumn{1}{c}{2024} & \multicolumn{1}{c}{2024} & \multicolumn{1}{c}{2024} & \multicolumn{1}{c}{2025} & \multicolumn{1}{c}{2025} & \multicolumn{1}{c|}{2025} &  \\
\hline
\multicolumn{2}{r|}{\multirow{4}{*}{Runtimes (s)}} & \multicolumn{1}{r|}{Houston} &\multicolumn{1}{c}{85.11} & \multicolumn{1}{c}{13.42} & \multicolumn{1}{c}{27.75} & \multicolumn{1}{c}{118.87} & \multicolumn{1}{c}{361.08} & \multicolumn{1}{c}{51.97} & \multicolumn{1}{c|}{9.59} & \multicolumn{1}{c}{33.12} \\
\multicolumn{2}{r|}{}& \multicolumn{1}{r|}{Moffett} &\multicolumn{1}{c}{67.14} & \multicolumn{1}{c}{11.06} & \multicolumn{1}{c}{14.25} & \multicolumn{1}{c}{112.73} & \multicolumn{1}{c}{72.97} & \multicolumn{1}{c}{41.78} & \multicolumn{1}{c|}{7.20} & \multicolumn{1}{c}{32.86} \\
\multicolumn{2}{r|}{}& \multicolumn{1}{r|}{Jasper-Ridge} &\multicolumn{1}{c}{86.27} & \multicolumn{1}{c}{13.30} & \multicolumn{1}{c}{28.81} & \multicolumn{1}{c}{118.55} & \multicolumn{1}{c}{328.77} & \multicolumn{1}{c}{51.67} & \multicolumn{1}{c|}{9.47} & \multicolumn{1}{c}{25.98} \\
\multicolumn{2}{r|}{}& \multicolumn{1}{r|}{Simulated} &\multicolumn{1}{c}{97.05} & \multicolumn{1}{c}{15.03} & \multicolumn{1}{c}{38.62} & \multicolumn{1}{c}{133.64} & \multicolumn{1}{c}{567.80} & \multicolumn{1}{c}{75.98} & \multicolumn{1}{c|}{11.64} & \multicolumn{1}{c}{50.12} \\
\hline
\multicolumn{1}{r|}{\multirow{17}{*}{SAD}} & \multicolumn{1}{r|}{\multirow{5}{*}{Houston}} & \multicolumn{1}{r|}{Grass} & 0.094\textbf{$\pm$}0.003 & \textcolor{c2}{\textbf{0.031}}\textbf{$\pm$}0.004 & 0.070\textbf{$\pm$}0.022 & 0.105\textbf{$\pm$}0.004 & 0.070\textbf{$\pm$}0.008 & 0.113\textbf{$\pm$}0.002 & 0.058\textbf{$\pm$}\textcolor{c2}{\textbf{0.001}} & \textcolor{c3}{\textbf{0.006}}\textbf{$\pm$}\textcolor{c3}{\textbf{0.000}} \\
&  & \multicolumn{1}{r|}{Runway} & 0.123\textbf{$\pm$}0.062 & 0.158\textbf{$\pm$}0.052 & 0.112\textbf{$\pm$}0.041 & 0.228\textbf{$\pm$}0.009 & 0.180\textbf{$\pm$}0.051 & 0.111\textbf{$\pm$}0.054 & \textcolor{c2}{\textbf{0.097}}\textbf{$\pm$}\textcolor{c2}{\textbf{0.003}} & \textcolor{c3}{\textbf{0.069}}\textbf{$\pm$}\textcolor{c3}{\textbf{0.001}} \\
&  & \multicolumn{1}{r|}{Parking lot 1} & 0.111\textbf{$\pm$}0.037 & 0.103\textbf{$\pm$}0.046 & 0.095\textbf{$\pm$}0.047 & 0.105\textbf{$\pm$}0.011 & 0.168\textbf{$\pm$}0.209 & 0.067\textbf{$\pm$}0.012 & \textcolor{c2}{\textbf{0.058}}\textbf{$\pm$}\textcolor{c2}{\textbf{0.001}} & \textcolor{c3}{\textbf{0.004}}\textbf{$\pm$}\textcolor{c3}{\textbf{0.000}} \\
&  & \multicolumn{1}{r|}{Parking lot 2} & 0.319\textbf{$\pm$}0.230 & 0.175\textbf{$\pm$}0.024 & \textcolor{c2}{\textbf{0.083}}\textbf{$\pm$}0.013 & 0.179\textbf{$\pm$}0.018 & 0.262\textbf{$\pm$}0.282 & 0.217\textbf{$\pm$}\textcolor{c2}{\textbf{0.007}} & 0.087\textbf{$\pm$}0.009 & \textcolor{c3}{\textbf{0.036}}\textbf{$\pm$}\textcolor{c3}{\textbf{0.000}} \\
&  & \multicolumn{1}{r|}{Average} & 0.162\textbf{$\pm$}0.067 & 0.117\textbf{$\pm$}0.020 & 0.090\textbf{$\pm$}0.016 & 0.154\textbf{$\pm$}0.006 & 0.170\textbf{$\pm$}0.131 & 0.127\textbf{$\pm$}0.019 & \textcolor{c2}{\textbf{0.075}}\textbf{$\pm$}\textcolor{c2}{\textbf{0.003}} & \textcolor{c3}{\textbf{0.028}}\textbf{$\pm$}\textcolor{c3}{\textbf{0.000}} \\
\cmidrule{2-11}
& \multicolumn{1}{r|}{\multirow{4}{*}{Moffett}} & \multicolumn{1}{r|}{Water} & 0.029\textbf{$\pm$}0.009 & 0.038\textbf{$\pm$}0.012 & 0.079\textbf{$\pm$}0.032 & 0.119\textbf{$\pm$}0.026 & \textcolor{c2}{\textbf{0.021}}\textbf{$\pm$}\textcolor{c3}{\textbf{0.000}} & 0.114\textbf{$\pm$}0.066 & 0.022\textbf{$\pm$}\textcolor{c2}{\textbf{0.005}} & \textcolor{c3}{\textbf{0.017}}\textbf{$\pm$}\textcolor{c3}{\textbf{0.000}} \\
&  & \multicolumn{1}{r|}{Soil} & 0.022\textbf{$\pm$}0.004 & 0.072\textbf{$\pm$}0.036 & 0.176\textbf{$\pm$}0.152 & 0.061\textbf{$\pm$}0.115 & \textcolor{c3}{\textbf{0.009}}\textbf{$\pm$}\textcolor{c2}{\textbf{0.001}} & 0.021\textbf{$\pm$}0.009 & \textcolor{c2}{\textbf{0.014}}\textbf{$\pm$}0.004 & 0.015\textbf{$\pm$}\textcolor{c3}{\textbf{0.000}} \\
&  & \multicolumn{1}{r|}{Vegetation} & 0.202\textbf{$\pm$}0.091 & 0.420\textbf{$\pm$}0.288 & 0.115\textbf{$\pm$}0.017 & 0.052\textbf{$\pm$}0.031 & 0.086\textbf{$\pm$}0.017 & 0.105\textbf{$\pm$}0.046 & \textcolor{c3}{\textbf{0.024}}\textbf{$\pm$}\textcolor{c2}{\textbf{0.009}} & \textcolor{c2}{\textbf{0.029}}\textbf{$\pm$}\textcolor{c3}{\textbf{0.001}} \\
&  & \multicolumn{1}{r|}{Average} & 0.084\textbf{$\pm$}0.030 & 0.177\textbf{$\pm$}0.108 & 0.123\textbf{$\pm$}0.061 & 0.077\textbf{$\pm$}0.042 & \textcolor{c2}{\textbf{0.039}}\textbf{$\pm$}\textcolor{c2}{\textbf{0.005}} & 0.080\textbf{$\pm$}0.034 & \textcolor{c3}{\textbf{0.020}}\textbf{$\pm$}\textcolor{c2}{\textbf{0.005}} & \textcolor{c3}{\textbf{0.020}}\textbf{$\pm$}\textcolor{c3}{\textbf{0.000}} \\
\cmidrule{2-11}
& \multicolumn{1}{r|}{\multirow{5}{*}{\makecell{Jasper-\\Ridge}}} & \multicolumn{1}{r|}{Vegetation} & 0.073\textbf{$\pm$}0.022 & 0.104\textbf{$\pm$}0.008 & 0.152\textbf{$\pm$}0.018 & \textcolor{c2}{\textbf{0.058}}\textbf{$\pm$}\textcolor{c2}{\textbf{0.001}} & 0.098\textbf{$\pm$}0.023 & \textcolor{c3}{\textbf{0.040}}\textbf{$\pm$}0.002 & 0.152\textbf{$\pm$}0.002 & 0.101\textbf{$\pm$}\textcolor{c3}{\textbf{0.000}} \\
&  & \multicolumn{1}{r|}{Water} & \textcolor{c2}{\textbf{0.048}}\textbf{$\pm$}0.015 & 0.134\textbf{$\pm$}\textcolor{c3}{\textbf{0.003}} & 0.096\textbf{$\pm$}0.042 & 0.218\textbf{$\pm$}\textcolor{c2}{\textbf{0.004}} & 0.087\textbf{$\pm$}0.028 & 0.240\textbf{$\pm$}0.005 & 0.069\textbf{$\pm$}0.015 & \textcolor{c3}{\textbf{0.046}}\textbf{$\pm$}0.007 \\
&  & \multicolumn{1}{r|}{Soil} & 0.135\textbf{$\pm$}0.025 & 0.098\textbf{$\pm$}0.047 & 0.171\textbf{$\pm$}0.031 & 0.136\textbf{$\pm$}0.007 & 0.072\textbf{$\pm$}0.015 & \textcolor{c3}{\textbf{0.022}}\textbf{$\pm$}\textcolor{c2}{\textbf{0.002}} & 0.104\textbf{$\pm$}0.010 & \textcolor{c2}{\textbf{0.067}}\textbf{$\pm$}\textcolor{c3}{\textbf{0.000}} \\
&  & \multicolumn{1}{r|}{Road} & 0.338\textbf{$\pm$}0.157 & 0.189\textbf{$\pm$}0.061 & 0.201\textbf{$\pm$}0.042 & 0.321\textbf{$\pm$}0.088 & 0.473\textbf{$\pm$}0.440 & \textcolor{c3}{\textbf{0.025}}\textbf{$\pm$}\textcolor{c3}{\textbf{0.002}} & 0.054\textbf{$\pm$}\textcolor{c2}{\textbf{0.014}} & \textcolor{c2}{\textbf{0.031}}\textbf{$\pm$}\textcolor{c3}{\textbf{0.002}} \\
&  & \multicolumn{1}{r|}{Average} & 0.149\textbf{$\pm$}0.047 & 0.131\textbf{$\pm$}0.012 & 0.155\textbf{$\pm$}0.007 & 0.183\textbf{$\pm$}0.020 & 0.183\textbf{$\pm$}0.110 & \textcolor{c2}{\textbf{0.082}}\textbf{$\pm$}\textcolor{c3}{\textbf{0.001}} & 0.095\textbf{$\pm$}0.008 & \textcolor{c3}{\textbf{0.061}}\textbf{$\pm$}\textcolor{c2}{\textbf{0.002}} \\
\cmidrule{2-11}
& \multicolumn{1}{r|}{\multirow{3}{*}{Simulated}} & SNR=10 dB & 0.274\textbf{$\pm$}0.041 & 0.151\textbf{$\pm$}0.023 & \textcolor{c2}{\textbf{0.112}}\textbf{$\pm$}0.024 & 0.129\textbf{$\pm$}0.015 & 0.476\textbf{$\pm$}0.294 & 0.153\textbf{$\pm$}\textcolor{c2}{\textbf{0.008}} & 0.216\textbf{$\pm$}0.063 & \textcolor{c3}{\textbf{0.056}}\textbf{$\pm$}\textcolor{c3}{\textbf{0.001}} \\
&  & SNR=20 dB & 0.131\textbf{$\pm$}0.008 & 0.139\textbf{$\pm$}0.016 & 0.115\textbf{$\pm$}0.024 & 0.142\textbf{$\pm$}\textcolor{c2}{\textbf{0.006}} & 0.366\textbf{$\pm$}0.123 & 0.166\textbf{$\pm$}\textcolor{c3}{\textbf{0.001}} & \textcolor{c2}{\textbf{0.072}}\textbf{$\pm$}0.029 & \textcolor{c3}{\textbf{0.015}}\textbf{$\pm$}\textcolor{c3}{\textbf{0.001}} \\
&  & SNR=30 dB & \textcolor{c2}{\textbf{0.094}}\textbf{$\pm$}0.023 & 0.154\textbf{$\pm$}0.029 & 0.121\textbf{$\pm$}0.009 & 0.147\textbf{$\pm$}0.003 & 0.428\textbf{$\pm$}0.300 & 0.155\textbf{$\pm$}\textcolor{c2}{\textbf{0.002}} & 0.100\textbf{$\pm$}0.006 & \textcolor{c3}{\textbf{0.018}}\textbf{$\pm$}\textcolor{c3}{\textbf{0.001}} \\
\hline
\multicolumn{1}{r|}{\multirow{17}{*}{RMSE}} & \multicolumn{1}{r|}{\multirow{5}{*}{Houston}} & \multicolumn{1}{r|}{Grass} & 0.096\textbf{$\pm$}0.023 & \textcolor{c2}{\textbf{0.056}}\textbf{$\pm$}0.014 & 0.066\textbf{$\pm$}0.013 & 0.074\textbf{$\pm$}0.005 & 0.147\textbf{$\pm$}0.132 & 0.071\textbf{$\pm$}0.005 & \textcolor{c3}{\textbf{0.041}}\textbf{$\pm$}\textcolor{c2}{\textbf{0.003}} & 0.067\textbf{$\pm$}\textcolor{c3}{\textbf{0.001}} \\
&  & \multicolumn{1}{r|}{Runway} & 0.102\textbf{$\pm$}0.038 & 0.125\textbf{$\pm$}0.044 & 0.161\textbf{$\pm$}0.052 & 0.258\textbf{$\pm$}0.051 & 0.184\textbf{$\pm$}0.017 & 0.119\textbf{$\pm$}0.091 & \textcolor{c3}{\textbf{0.043}}\textbf{$\pm$}\textcolor{c2}{\textbf{0.003}} & \textcolor{c2}{\textbf{0.047}}\textbf{$\pm$}\textcolor{c3}{\textbf{0.002}} \\
&  & Parking lot 1 & 0.235\textbf{$\pm$}0.044 & 0.145\textbf{$\pm$}0.040 & 0.209\textbf{$\pm$}0.038 & 0.309\textbf{$\pm$}0.029 & 0.244\textbf{$\pm$}\textcolor{c2}{\textbf{0.007}} & 0.294\textbf{$\pm$}0.019 & \textcolor{c2}{\textbf{0.115}}\textbf{$\pm$}0.010 & \textcolor{c3}{\textbf{0.062}}\textbf{$\pm$}\textcolor{c3}{\textbf{0.001}} \\
&  & Parking lot 2 & 0.209\textbf{$\pm$}0.054 & \textcolor{c2}{\textbf{0.092}}\textbf{$\pm$}0.041 & 0.197\textbf{$\pm$}0.056 & 0.278\textbf{$\pm$}\textcolor{c2}{\textbf{0.009}} & 0.256\textbf{$\pm$}0.177 & 0.267\textbf{$\pm$}0.037 & 0.103\textbf{$\pm$}\textcolor{c2}{\textbf{0.009}} & \textcolor{c3}{\textbf{0.086}}\textbf{$\pm$}\textcolor{c3}{\textbf{0.002}} \\
&  & \multicolumn{1}{r|}{Average} & 0.161\textbf{$\pm$}0.025 & 0.105\textbf{$\pm$}0.027 & 0.158\textbf{$\pm$}0.020 & 0.230\textbf{$\pm$}0.006 & 0.208\textbf{$\pm$}0.077 & 0.188\textbf{$\pm$}0.034 & \textcolor{c2}{\textbf{0.076}}\textbf{$\pm$}\textcolor{c2}{\textbf{0.004}} & \textcolor{c3}{\textbf{0.066}}\textbf{$\pm$}\textcolor{c3}{\textbf{0.001}} \\
\cmidrule{2-11}
& \multicolumn{1}{r|}{\multirow{4}{*}{Moffett}} & \multicolumn{1}{r|}{Water} & \textcolor{c2}{\textbf{0.053}}\textbf{$\pm$}\textcolor{c2}{\textbf{0.007}} & 0.200\textbf{$\pm$}0.101 & 0.196\textbf{$\pm$}0.054 & 0.094\textbf{$\pm$}0.076 & 0.065\textbf{$\pm$}\textcolor{c3}{\textbf{0.003}} & \textcolor{c3}{\textbf{0.039}}\textbf{$\pm$}0.020 & 0.073\textbf{$\pm$}\textcolor{c2}{\textbf{0.007}} & 0.061\textbf{$\pm$}0.011 \\
&  & \multicolumn{1}{r|}{Soil} & \textcolor{c2}{\textbf{0.056}}\textbf{$\pm$}0.008 & 0.209\textbf{$\pm$}0.105 & 0.151\textbf{$\pm$}0.086 & 0.082\textbf{$\pm$}0.065 & 0.077\textbf{$\pm$}\textcolor{c2}{\textbf{0.007}} & 0.068\textbf{$\pm$}0.011 & \textcolor{c2}{\textbf{0.056}}\textbf{$\pm$}0.008 & \textcolor{c3}{\textbf{0.051}}\textbf{$\pm$}\textcolor{c3}{\textbf{0.004}}\\
&  & \multicolumn{1}{r|}{Vegetation} & 0.067\textbf{$\pm$}0.013 & 0.232\textbf{$\pm$}0.131 & 0.211\textbf{$\pm$}0.086 & 0.088\textbf{$\pm$}0.090 & 0.086\textbf{$\pm$}0.016 & 0.073\textbf{$\pm$}\textcolor{c2}{\textbf{0.011}} & \textcolor{c2}{\textbf{0.052}}\textbf{$\pm$}\textcolor{c3}{\textbf{0.005}} & \textcolor{c3}{\textbf{0.041}}\textbf{$\pm$}\textcolor{c3}{\textbf{0.005}}\\
&  & \multicolumn{1}{r|}{Average} & \textcolor{c2}{\textbf{0.059}}\textbf{$\pm$}0.007 & 0.214\textbf{$\pm$}0.109 & 0.186\textbf{$\pm$}0.074 & 0.088\textbf{$\pm$}0.077 & 0.076\textbf{$\pm$}0.008 & 0.060\textbf{$\pm$}0.012 & 0.060\textbf{$\pm$}\textcolor{c3}{\textbf{0.005}} & \textcolor{c3}{\textbf{0.051}}\textbf{$\pm$}\textcolor{c2}{\textbf{0.006}} \\
\cmidrule{2-11}
& \multicolumn{1}{r|}{\multirow{5}{*}{\makecell{Jasper-\\Ridge}}} & \multicolumn{1}{r|}{Vegetation} & 0.141\textbf{$\pm$}0.052 & 0.123\textbf{$\pm$}0.023 & 0.138\textbf{$\pm$}0.031 & 0.104\textbf{$\pm$}\textcolor{c2}{\textbf{0.003}} & \textcolor{c2}{\textbf{0.078}}\textbf{$\pm$}0.017 & 0.092\textbf{$\pm$}0.004 & 0.126\textbf{$\pm$}0.016 & \textcolor{c3}{\textbf{0.065}}\textbf{$\pm$}\textcolor{c3}{\textbf{0.001}} \\
&  & \multicolumn{1}{r|}{Water} & \textcolor{c3}{\textbf{0.053}}\textbf{$\pm$}0.012 & 0.097\textbf{$\pm$}\textcolor{c3}{\textbf{0.001}} & 0.073\textbf{$\pm$}0.016 & 0.125\textbf{$\pm$}0.006 & 0.073\textbf{$\pm$}0.011 & 0.114\textbf{$\pm$}\textcolor{c2}{\textbf{0.003}} & \textcolor{c2}{\textbf{0.061}}\textbf{$\pm$}0.009 & 0.069\textbf{$\pm$}0.004 \\
&  & \multicolumn{1}{r|}{Soil} & 0.215\textbf{$\pm$}0.029 & 0.139\textbf{$\pm$}0.023 & 0.163\textbf{$\pm$}0.026 & 0.225\textbf{$\pm$}0.015 & 0.217\textbf{$\pm$}0.021 & \textcolor{c3}{\textbf{0.085}}\textbf{$\pm$}\textcolor{c3}{\textbf{0.004}} & 0.129\textbf{$\pm$}0.011 & \textcolor{c2}{\textbf{0.107}}\textbf{$\pm$}\textcolor{c2}{\textbf{0.006}} \\
&  & \multicolumn{1}{r|}{Road} & 0.247\textbf{$\pm$}0.079 & 0.116\textbf{$\pm$}0.027 & 0.159\textbf{$\pm$}0.037 & 0.216\textbf{$\pm$}0.012 & 0.217\textbf{$\pm$}0.012 & \textcolor{c3}{\textbf{0.070}}\textbf{$\pm$}\textcolor{c3}{\textbf{0.003}} & \textcolor{c2}{\textbf{0.084}}\textbf{$\pm$}\textcolor{c2}{\textbf{0.007}} & 0.101\textbf{$\pm$}0.009 \\
&  & \multicolumn{1}{r|}{Average} & 0.164\textbf{$\pm$}0.035 & 0.119\textbf{$\pm$}0.017 & 0.133\textbf{$\pm$}0.019 & 0.168\textbf{$\pm$}0.009 & 0.146\textbf{$\pm$}0.008 & \textcolor{c2}{\textbf{0.090}}\textbf{$\pm$}\textcolor{c3}{\textbf{0.003}} & 0.100\textbf{$\pm$}0.008 & \textcolor{c3}{\textbf{0.086}}\textbf{$\pm$}\textcolor{c2}{\textbf{0.004}} \\
\cmidrule{2-11}
& \multicolumn{1}{r|}{\multirow{3}{*}{Simulated}} & SNR=10 dB & 0.215\textbf{$\pm$}0.035 & 0.225\textbf{$\pm$}0.032 & \textcolor{c2}{\textbf{0.190}}\textbf{$\pm$}0.027 & 0.239\textbf{$\pm$}0.063 & 0.381\textbf{$\pm$}0.044 & 0.321\textbf{$\pm$}\textcolor{c2}{\textbf{0.018}} & 0.211\textbf{$\pm$}\textcolor{c3}{\textbf{0.006}} & \textcolor{c3}{\textbf{0.121}}\textbf{$\pm$}\textcolor{c3}{\textbf{0.006}} \\
&  & SNR=20 dB & 0.190\textbf{$\pm$}0.015 & 0.262\textbf{$\pm$}0.032 & 0.226\textbf{$\pm$}0.035 & 0.227\textbf{$\pm$}0.024 & 0.341\textbf{$\pm$}0.015 & 0.227\textbf{$\pm$}\textcolor{c3}{\textbf{0.003}} & \textcolor{c2}{\textbf{0.122}}\textbf{$\pm$}0.034 & \textcolor{c3}{\textbf{0.083}}\textbf{$\pm$}\textcolor{c2}{\textbf{0.005}} \\
&  & SNR=30 dB & \textcolor{c2}{\textbf{0.144}}\textbf{$\pm$}0.029 & 0.219\textbf{$\pm$}0.015 & 0.171\textbf{$\pm$}0.021 & 0.233\textbf{$\pm$}\textcolor{c2}{\textbf{0.006}} & 0.360\textbf{$\pm$}0.049 & 0.262\textbf{$\pm$}0.030 & 0.150\textbf{$\pm$}0.019 & \textcolor{c3}{\textbf{0.077}}\textbf{$\pm$}\textcolor{c3}{\textbf{0.003}} \\
\hline
\end{tabular}
\label{tab:sota}
\end{table*}
\subsection{Dataset Description}
\label{exp_datasets}
\begin{figure}[t!]
	\centering
	\includegraphics[width=0.49\textwidth]{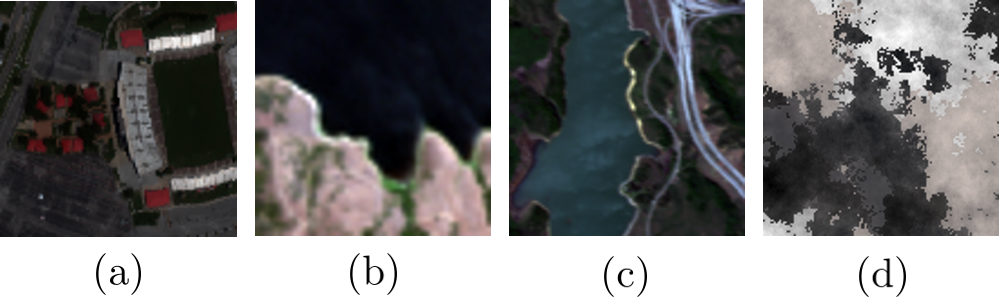}
	\caption{Visualization of datasets: (a) Houston, (b) Moffett, (c) Jasper-Ridge, (d) Simulated.}
	\label{fig:datasets}
\end{figure}

We used three real-world datasets and one simulated dataset, as shown in Figure \ref{fig:datasets}, to evaluate our proposed method. The datasets are described below.

\noindent \textbf{Houston Dataset \cite{houston}: }This dataset was collected by the ITRES CASI-1500 sensor on the University of Houston campus in Houston, Texas, USA, and was initially released
by the IEEE GRSS Data Fusion Contest in 2013. The original image consists of
$394 \times 1905$ pixels and spans a spectral range from $364$ to $1046$ nm across $144$ bands. For this study, we selected a $105 \times 105$ pixel subimage cropped from the original, as presented in Figure \ref{fig:datasets}-(a). The chosen area primarily includes four endmembers: grass, runway, parking lot 1, and parking lot 2.

\noindent \textbf{Moffett Dataset \cite{moffett}: }The Moffett
dataset is collected by the AVIRIS
sensor in the Moffett area at the southern end of San Francisco, California, USA. This dataset, presented in Figure \ref{fig:datasets}-(b), covers a wavelength range
from $400$ to $2500$ nm, and each pixel contains $184$ bands. The dataset contains $50 \times 50$ pixels, containing three objects: soil, vegetation, and water.  

\noindent \textbf{Jasper-Ridge Dataset \cite{jasper_ridge}: }The Jasper
Ridge data was collected by the AVIRIS sensor in central California, United States. The dataset, presented in Figure \ref{fig:datasets}-(c), comprises $100\times 100$ pixels, and the scenario contains four materials: water, soil, tree, and road. The original data has $224$ bands, spanning a spectral range from $380$ to $2500$ nm. In this experiment,
we retained $198$ bands because some bands were removed to reduce the effects of significant water vapor and atmospheric interferences. 

\noindent \textbf{Simulated Dataset: }To
evaluate the HSU performance under controlled conditions, we generate simulated data. Five endmembers’ spectra for a simulated
dataset with 180 bands are extracted in the USGS digital spectral library\footnote[1]{Available online: \url{https://www.usgs.gov/labs/spectroscopy-lab/science/spectral-library}}. The abundance maps with the spatial size $130 \times 130$ are generated through the hyperspectral imagery synthesis (HYDRA) toolbox\footnote[2]{Available online: \url{https://www.ehu.eus/ccwintco/index.php/Hyperspectral_Imagery_Synthesis_tools_for_MATLAB}} through the Gaussian field method. The simulated dataset is shown in Figure \ref{fig:datasets}-(d). We added zero mean Gaussian noise with three SNR levels: 10dB, 20dB, and 30dB.

\begin{figure*}[htbp] 
	\centering
	\begin{tabular}{cc} 
		
		\includegraphics[width=\textwidth]{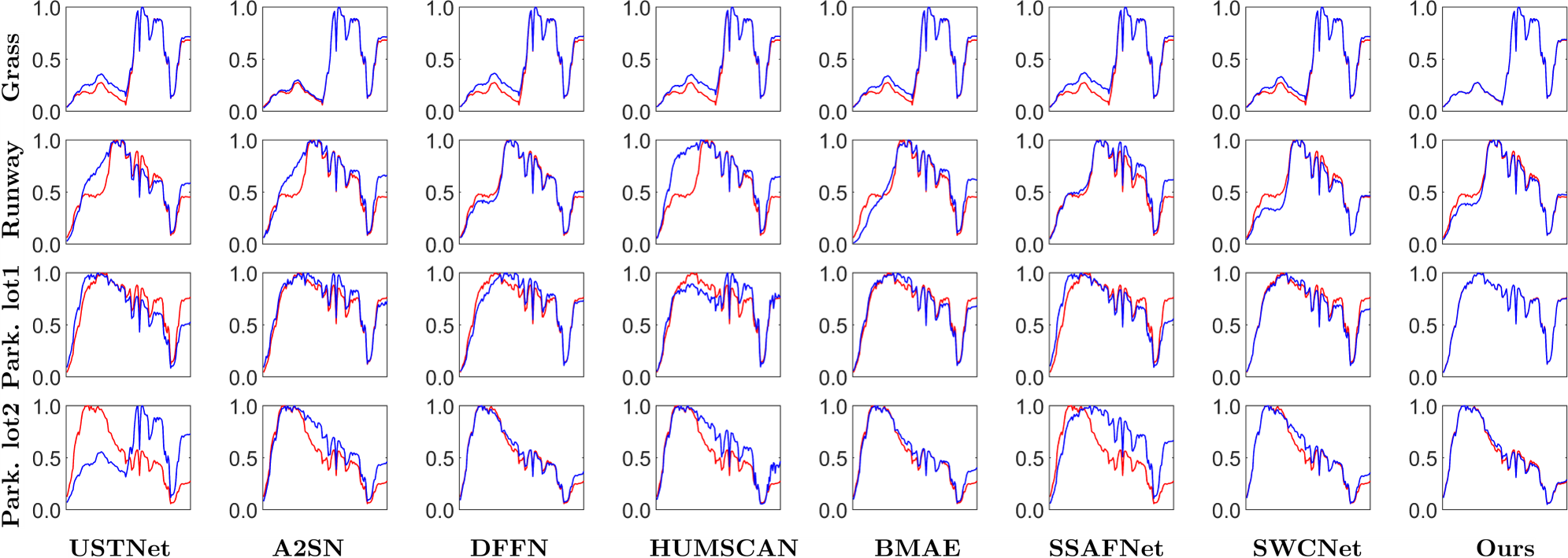} \\
		[4pt]
		(a) Estimated endmembers using different HSU methods. Blue line: extracted endmember. Red line: reference GT. \\
		 [10pt]
		\includegraphics[width=\textwidth]{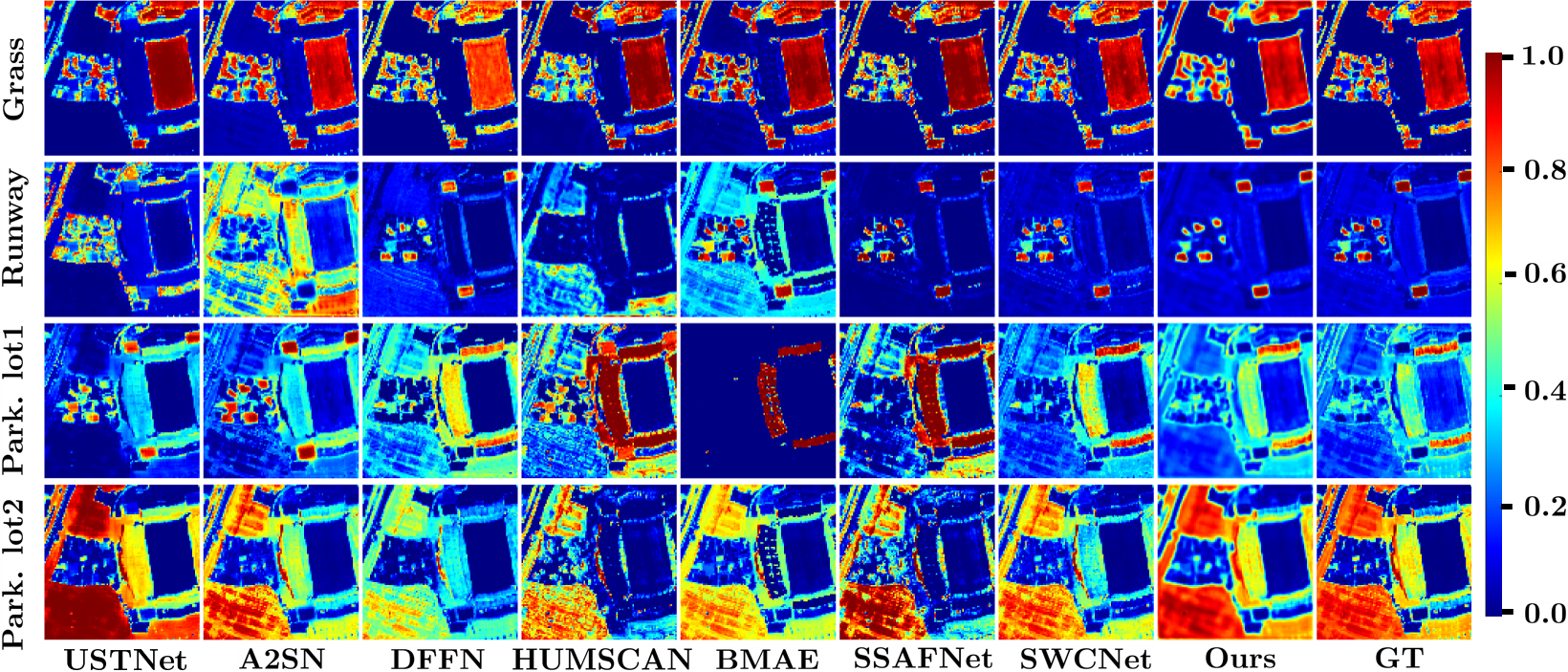} \\ 
		[4pt]
		 (b) Estimated abundance maps using different HSU methods. \\ 
	\end{tabular}
	\caption{Visualization on Houston dataset: (a) Estimated endmembers, (b) Abundance maps.} 
	\label{fig:sota_houston} 
\end{figure*}
\begin{table}[t!]
	\fontsize{7.4}{8}\selectfont
	\centering
	\caption{Hyperparameters settings for all datasets.} 
	\begin{tabular}{rcccc}
		\hline
		\multicolumn{1}{r}{Hyperparameters} & \multicolumn{1}{c}{$L_E$} & \multicolumn{1}{c}{$L_D$} & \multicolumn{1}{c}{$T_1$} & \multicolumn{1}{c}{$T$} \\
		
		\hline
		Houston & 1e-4&1e-5&500&1000\\
		Moffett &2e-4&2e-4&1890&2000\\
		Jasper-Ridge &1.2e-4&1e-4&500&2000\\
		Simulated &1.2e-4&1e-4&900&1000\\
		\hline
	\end{tabular}
	\label{tab:hyperparameters}
\end{table}
\subsection{Experimental Setup}
\label{exp_setup}
\noindent \textbf{Evaluation metrics: }To compare the HSU performance, we use two quantitative metrics: spectral angle distance (SAD) and root mean square error (RMSE). For the $i^{th}$ material, the mathematical expressions of SAD and RMSE are,

\begin{equation}\label{eq:sad}
	\text{SAD}(a_i,\hat{a}_i) = \text{cos}^{-1}\Bigg(\frac{<e_i, \hat{e}_i>}{||e_i||_2||\hat{e}_i||_2}\Bigg)
\end{equation}

\begin{equation}\label{eq:rmse}
	\text{RMSE}(a_i,\hat{a}_i) = \sqrt{\frac{1}{N}\sum_{i=1}^{N}||a_i-\hat{a}_i||_2^2}
\end{equation}

\noindent where $e_i$ and $\hat{e}_i$ are the estimated and ground-truth (GT) endmembers, and $a_i$ and $\hat{a}_i$ are the estimated and ground-truth (GT) abundances. A lower value of SAD and RMSE is desired.

\noindent \textbf{Training settings: } We train 3D-CSCNet by minimizing the loss function in Eqn. \ref{eq:loss}. The training is performed with Adam optimizer for a total number of epochs $T$. In the training stage-I, we train only the encoder with learning rate $L_E$ for $T_1$ number of epochs. In the training stage-II, we train the encoder and with learning rate $L_E$, and the decoder and with learning rate $L_D$ for $T-T_1$ number of epochs. For the four datasets, the values of the hyperparameters $L_E,L_D,T_1,T$ are tabulated in Table \ref{tab:hyperparameters}. We have conducted all experiments using an NVIDIA A40 GPU within the Pytorch framework.

\noindent \textbf{Implementation details: } In the encoder of 3D-CSCNet, each convolution layer is set to have a number of filters $C=48$. The kernel sizes for different convolution layers are,  $W_{u}^{(k)}(\cdot): 7\times 3\times 3,\:\:W_{d}^{(k)}(\cdot): 7\times 3\times 3,\:\:W_{in}^{(k)}(\cdot): 15\times 3\times 3,\:\: G(\cdot): 1\times 1\times 1$. We use six IMs in 3D-CSCB.

\begin{figure*}[htbp] 
	\centering
	\begin{tabular}{cc} 
		\includegraphics[width=\textwidth]{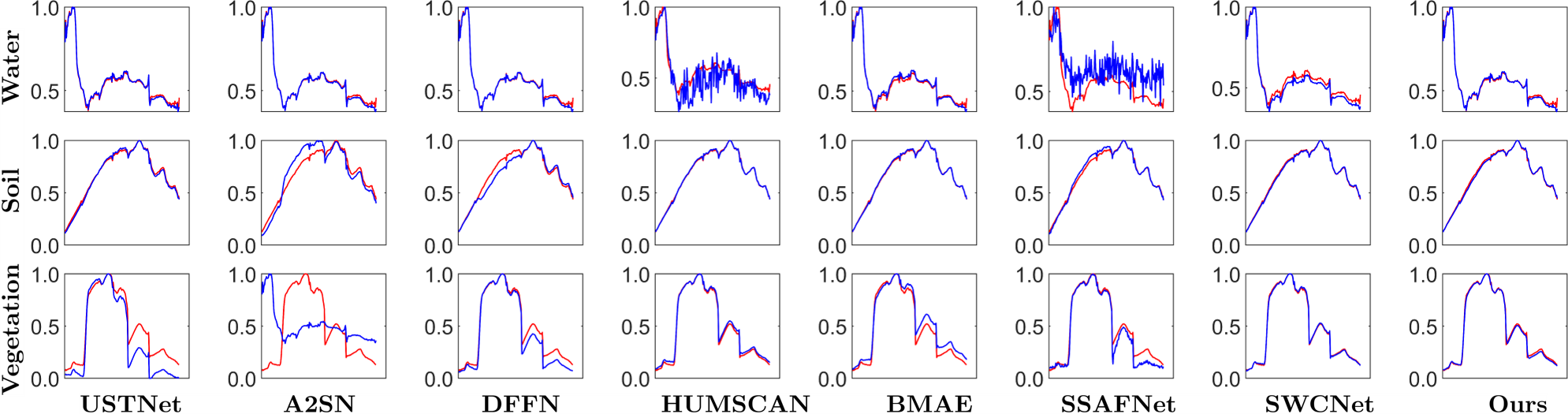} \\
		[4pt]
		(a) Estimated endmembers using different HSU methods. Blue line: extracted endmember. Red line: reference GT. \\
		[10pt]
		\includegraphics[width=\textwidth]{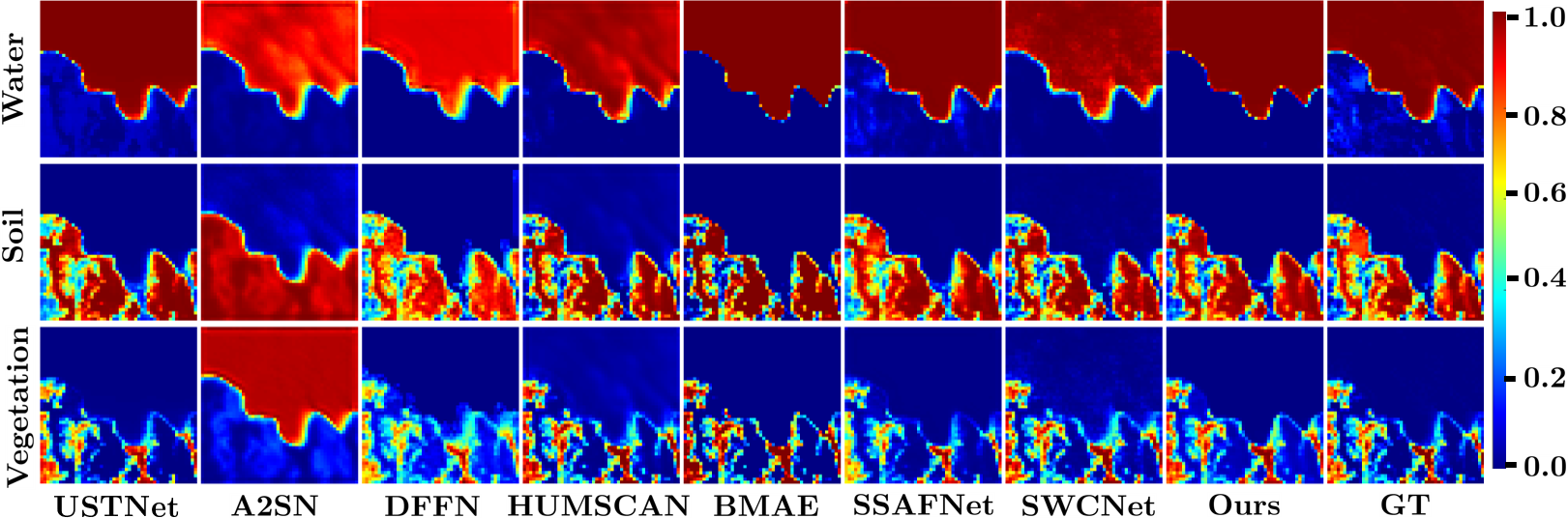} \\ 
		[4pt]
		(b) Estimated abundance maps using different HSU methods. \\ 
	\end{tabular}
	\caption{Visualization on Moffett dataset: (a) Estimated endmembers, (b) Abundance maps.} 
	\label{fig:sota_moffett} 
\end{figure*}
\begin{figure*}[htbp] 
	\centering
	\begin{tabular}{cc} 
		
		\includegraphics[width=\textwidth]{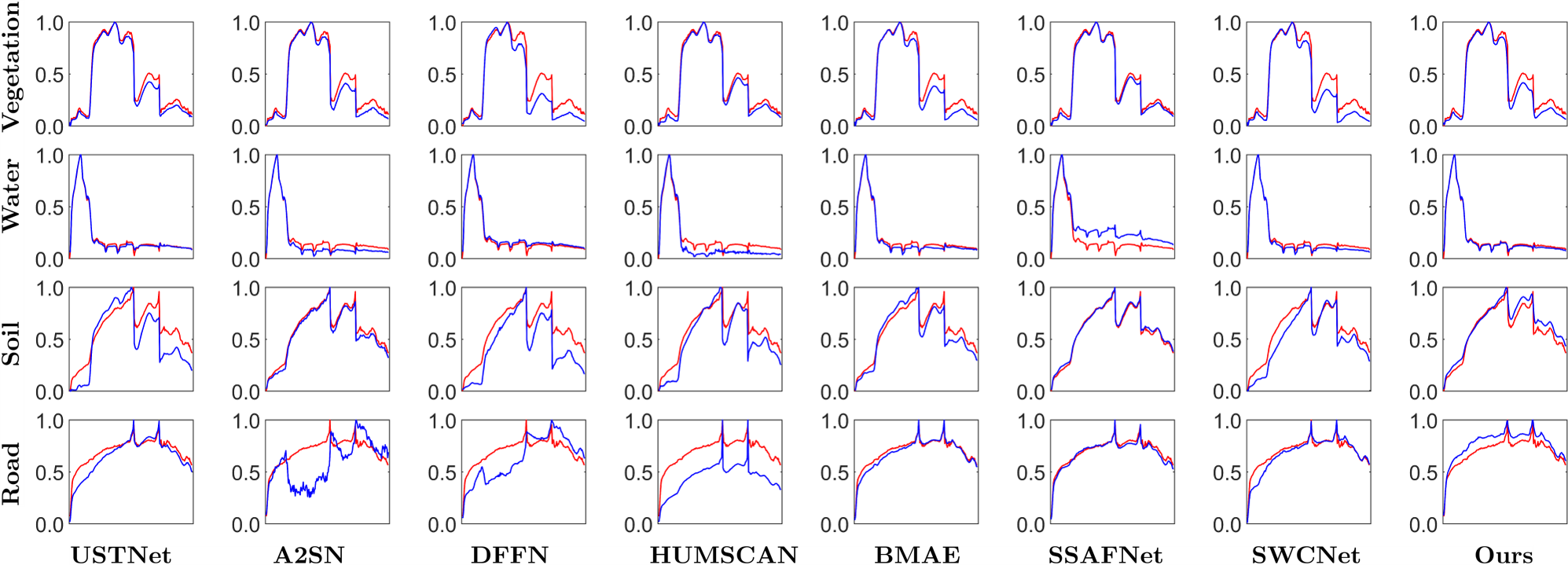} \\
		[4pt]
		(a) Estimated endmembers using different HSU methods. Blue line: extracted endmember. Red line: reference GT. \\
		[10pt]
		\includegraphics[width=\textwidth]{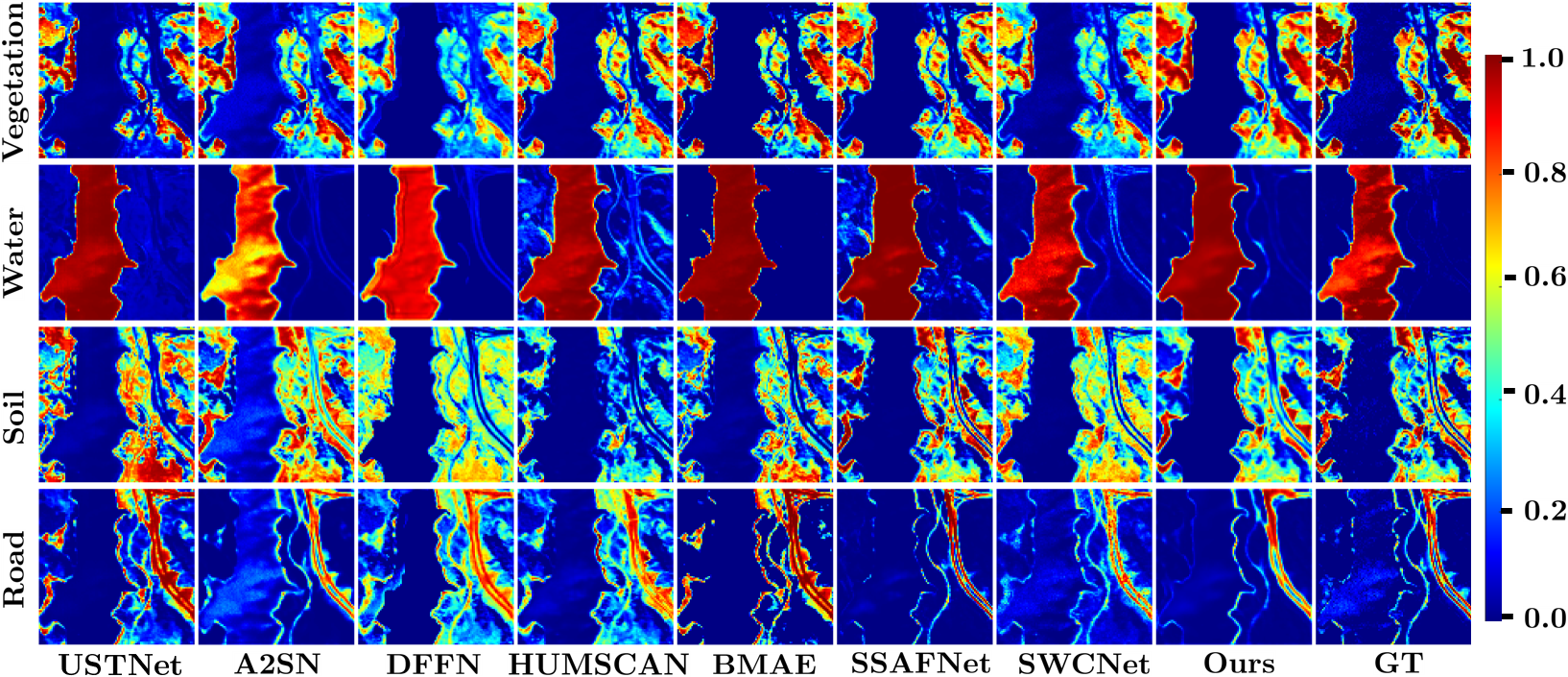} \\ 
		[4pt]
		(b) Estimated abundance maps using different HSU methods. \\ 
	\end{tabular}
	\caption{Visualization on Jasper-Ridge dataset: (a) Estimated endmembers, (b) Abundance maps.} 
	\label{fig:sota_jasper} 
\end{figure*}
\begin{table}[t!]
	\fontsize{7.4}{8}\selectfont
	\centering
	\caption{Comparison of SAD values using different endmember extraction algorithms.} 
	\begin{tabular}{B|P|FGB|G}
		\hline
		\multicolumn{2}{r|}{\multirow{1}{*}{Methods}} & \multicolumn{1}{c}{VCA} & \multicolumn{1}{c}{SVM} & \multicolumn{1}{c}{PSVM\textsubscript{ND}} & \multicolumn{1}{|c}{PSVM} \\
		
		\hline
		\multicolumn{1}{r|}{\multirow{5}{*}{\makecell{Houston \\ (SNR= \\ 36 dB)}}} & \multicolumn{1}{r|}{Grass} & \textcolor{c3}{\textbf{0.017}}\textbf{$\pm$}0.012 & 0.219 & \textcolor{c2}{\textbf{0.027}} & \textcolor{c2}{\textbf{0.027}} \\
		& \multicolumn{1}{r|}{Runway} & 0.138\textbf{$\pm$}0.093 & \textcolor{c2}{\textbf{0.086}} & \textcolor{c3}{\textbf{0.067}} & \textcolor{c3}{\textbf{0.067}} \\
		& \multicolumn{1}{r|}{Parking lot 1} & 0.057\textbf{$\pm$}0.040 & \textcolor{c2}{\textbf{0.046}} & \textcolor{c3}{\textbf{0.002}} & \textcolor{c3}{\textbf{0.002}} \\
		& \multicolumn{1}{r|}{Parking lot 2} & 0.123\textbf{$\pm$}0.241 & \textcolor{c2}{\textbf{0.057}} & \textcolor{c3}{\textbf{0.003}} & \textcolor{c3}{\textbf{0.003}} \\
		& \multicolumn{1}{r|}{Average} & \textcolor{c2}{\textbf{0.084}}\textbf{$\pm$}0.076 & 0.102 & \textcolor{c3}{\textbf{0.025}} & \textcolor{c3}{\textbf{0.025}} \\
		\hline
		\multicolumn{1}{r|}{\multirow{4}{*}{\makecell{Moffett \\ (SNR= \\ 33 dB)}}} & \multicolumn{1}{r|}{Water} & 0.143\textbf{$\pm$}0.000 & 0.130 & \textcolor{c2}{\textbf{0.098}} & \textcolor{c3}{\textbf{0.013}} \\
		& \multicolumn{1}{r|}{Soil} & 0.160\textbf{$\pm$}0.081 & 0.230 & \textcolor{c3}{\textbf{0.046}} & \textcolor{c2}{\textbf{0.075}}\\
		& \multicolumn{1}{r|}{Vegetation} & 0.105\textbf{$\pm$}0.080 & \textcolor{c3}{\textbf{0.020}} & 0.197 & \textcolor{c2}{\textbf{0.034}}\\
		& \multicolumn{1}{r|}{Average} & 0.136\textbf{$\pm$}0.001 & 0.127 & \textcolor{c2}{\textbf{0.114}} & \textcolor{c3}{\textbf{0.041}}\\
		\hline
		\multicolumn{1}{r|}{\multirow{5}{*}{\makecell{Jasper-\\Ridge \\ (SNR=\\30 dB)}}} & \multicolumn{1}{r|}{Vegetation} & \textcolor{c2}{\textbf{0.225}}\textbf{$\pm$}0.027 & 0.254 & 0.262 & \textcolor{c3}{\textbf{0.194}}\\
		& \multicolumn{1}{r|}{Water} & 0.243\textbf{$\pm$}0.065 & 0.107 & \textcolor{c2}{\textbf{0.090}} & \textcolor{c3}{\textbf{0.039}}\\
		& \multicolumn{1}{r|}{Soil} & 0.283\textbf{$\pm$}0.114 & 0.134 & \textcolor{c2}{\textbf{0.117}} & \textcolor{c3}{\textbf{0.069}}\\
		& \multicolumn{1}{r|}{Road} & 0.574\textbf{$\pm$}0.078 & 0.156 & \textcolor{c2}{\textbf{0.148}} & \textcolor{c3}{\textbf{0.100}}\\
		& \multicolumn{1}{r|}{Average} & 0.331\textbf{$\pm$}0.049 & 0.163 & \textcolor{c2}{\textbf{0.154}} & \textcolor{c3}{\textbf{0.101}}\\
		\hline
		\multicolumn{1}{r|}{\multirow{3}{*}{Simulated}} & SNR=10 dB & \textcolor{c2}{\textbf{0.185}}\textbf{$\pm$}0.038 & 0.651 & 0.193 & \textcolor{c3}{\textbf{0.059}}\\
		& SNR=20 dB & 0.040\textbf{$\pm$}0.005 & 0.163 & \textcolor{c2}{\textbf{0.018}} & \textcolor{c3}{\textbf{0.015}}\\
		& SNR=30 dB & \textcolor{c2}{\textbf{0.019}}\textbf{$\pm$}0.002 & 0.082 & 0.020 & \textcolor{c3}{\textbf{0.012}}\\
		\hline
	\end{tabular}
	\label{tab:algorithms}
\end{table}
\subsection{Performance Comparison with SOTA Methods}
\label{exp_sota}
We compare the performance of 3D-CSCNet with recent SOTA methods:  USTNet \cite{ustnet}, A2SN \cite{a2sn}, DFFN \cite{dffn}, HUMSCAN \cite{humscan}, BMAE \cite{bmae}, SSAFNet \cite{ssafnet}, and SWCNet \cite{swcnet}. For the SOTA methods, we use their official code to estimate the abundance and endmember matrices.

\noindent \textbf{Quantitative comparison: } Table \ref{tab:sota} shows the SAD and RMSE values on three real datasets, and one simulated dataset with three different SNR levels. The results are obtained after six different runs, and the metrics are presented as ``mean$\pm$standard deviation". A lower value of mean and standard deviation is desired. 3D-CSCNet has leading performance across the four datasets. Among the other SOTA methods, SSAFNet and SWCNet have comparable performance to 3D-CSCNet. Along with the SAD and RMSE values, we also list the GPU runtime of different methods for the four datasets. 3D-CSCNet has comparable runtime with the SOTA methods.

\noindent \textbf{Qualitative comparison: } Figure \ref{fig:sota_houston} shows the estimated endmembers and abundance maps on Houston dataset using different HSU methods. Compared to the SOTA methods, 3D-CSCNet better captures the patterns in the endmembers and abundance maps of grass, runway and two parking lot regions. Figure \ref{fig:sota_moffett} shows a visual comparison on Moffett dataset. Estimated endmembers and abundance maps of water, soil and vegetation regions using our 3D-CSCNet are more similar to the GTs compared to the SOTA methods. Figure \ref{fig:sota_jasper} shows the estimated endmembers and abundance maps of vegetation, water, soil, and road regions on Jasper-Ridge dataset using different HSU methods. Compared to the SOTA methods, 3D-CSCNet better captures the patterns in the endmembers and abundance maps.
\begin{table*}[t!]
	\fontsize{7.5}{8}\selectfont
	\centering
	\caption{Comparison of results on ablation experiments (AE). The best and second-best performances are highlighted in \textcolor{red}{\textbf{red}} and \textcolor{blue}{\textbf{blue}} colors, respectively.}
	\begin{tabular}{lFAFAFAFAFA}
		\hline
		& \multicolumn{2}{c}{Houston}   & \multicolumn{2}{c}{Moffett}   & \multicolumn{2}{c}{Jasper-Ridge}   & \multicolumn{2}{c}{SNR = 10 dB}  & \multicolumn{2}{c}{SNR = 20 dB} \\
		\cmidrule{2-11}
		& \multicolumn{1}{c}{mSAD} & \multicolumn{1}{c}{mRMSE} & \multicolumn{1}{c}{mSAD} & \multicolumn{1}{c}{mRMSE} & \multicolumn{1}{c}{mSAD} & \multicolumn{1}{c}{mRMSE} & \multicolumn{1}{c}{mSAD} & \multicolumn{1}{c}{mRMSE} & \multicolumn{1}{c}{mSAD} & \multicolumn{1}{c}{mRMSE} \\
		\hline
		AE1 & 0.035\textbf{$\pm$}\textcolor{c3}{\textbf{0.000}} & 0.085\textbf{$\pm$}0.003 & 0.032\textbf{$\pm$}0.023 & 0.099\textbf{$\pm$}0.060 & 0.090\textbf{$\pm$}0.010 & 0.111\textbf{$\pm$}0.034 & 0.059\textbf{$\pm$}\textcolor{c2}{\textbf{0.001}} & 0.138\textbf{$\pm$}0.039 & 0.018\textbf{$\pm$}\textcolor{c2}{\textbf{0.002}} & 0.094\textbf{$\pm$}0.024 \\
		AE2 & 0.032\textbf{$\pm$}\textcolor{c3}{\textbf{0.000}} & 0.072\textbf{$\pm$}\textcolor{c2}{\textbf{0.002}} & 0.025\textbf{$\pm$}\textcolor{c2}{\textbf{0.001}} & \textcolor{c2}{\textbf{0.058}}\textbf{$\pm$}\textcolor{c2}{\textbf{0.009}} & 0.101\textbf{$\pm$}\textcolor{c3}{\textbf{0.000}} & 0.086\textbf{$\pm$}\textcolor{c2}{\textbf{0.003}} & \textcolor{c2}{\textbf{0.057}}\textbf{$\pm$}\textcolor{c2}{\textbf{0.001}} & 0.137\textbf{$\pm$}0.037 & 0.019\textbf{$\pm$}\textcolor{c3}{\textbf{0.001}} & \textcolor{c2}{\textbf{0.086}}\textbf{$\pm$}\textcolor{c2}{\textbf{0.005}} \\
		AE3 & \textcolor{c2}{\textbf{0.029}}\textbf{$\pm$}\textcolor{c2}{\textbf{0.001}} & 0.077\textbf{$\pm$}0.012 & \textcolor{c2}{\textbf{0.021}}\textbf{$\pm$}\textcolor{c3}{\textbf{0.000}} & 0.070\textbf{$\pm$}0.016 & \textcolor{c2}{\textbf{0.067}}\textbf{$\pm$}0.004 & 0.117\textbf{$\pm$}0.026 & 0.060\textbf{$\pm$}0.002 & 0.202\textbf{$\pm$}0.038 & 0.019\textbf{$\pm$}\textcolor{c2}{\textbf{0.002}} & 0.134\textbf{$\pm$}0.025 \\
		AE4 & \textcolor{c2}{\textbf{0.029}}\textbf{$\pm$}\textcolor{c3}{\textbf{0.000}} & 0.073\textbf{$\pm$}\textcolor{c2}{\textbf{0.002}} & 0.028\textbf{$\pm$}0.003 & 0.085\textbf{$\pm$}0.018 & 0.094\textbf{$\pm$}0.004 & \textcolor{c2}{\textbf{0.084}}\textbf{$\pm$}\textcolor{c3}{\textbf{0.002}} & 0.061\textbf{$\pm$}\textcolor{c2}{\textbf{0.001}} & 0.214\textbf{$\pm$}\textcolor{c2}{\textbf{0.004}} & 0.022\textbf{$\pm$}\textcolor{c3}{\textbf{0.001}} & 0.180\textbf{$\pm$}0.020 \\
		AE5 & \textcolor{c3}{\textbf{0.028}}\textbf{$\pm$}\textcolor{c3}{\textbf{0.000}} & 0.072\textbf{$\pm$}\textcolor{c3}{\textbf{0.001}} & \textcolor{c2}{\textbf{0.021}}\textbf{$\pm$}\textcolor{c3}{\textbf{0.000}} & 0.066\textbf{$\pm$}0.010 & 0.069\textbf{$\pm$}0.013 & 0.091\textbf{$\pm$}\textcolor{c2}{\textbf{0.003}} & \textcolor{c3}{\textbf{0.056}}\textbf{$\pm$}\textcolor{c3}{\textbf{0.000}} & \textcolor{c3}{\textbf{0.121}}\textbf{$\pm$}\textcolor{c3}{\textbf{0.001}} & \textcolor{c3}{\textbf{0.015}}\textbf{$\pm$}\textcolor{c3}{\textbf{0.001}} & 0.088\textbf{$\pm$}0.011 \\
		AE6 & \textcolor{c3}{\textbf{0.028}}\textbf{$\pm$}\textcolor{c3}{\textbf{0.000}} & \textcolor{c3}{\textbf{0.065}}\textbf{$\pm$}\textcolor{c2}{\textbf{0.002}} & \textcolor{c2}{\textbf{0.021}}\textbf{$\pm$}\textcolor{c3}{\textbf{0.000}} & \textcolor{c2}{\textbf{0.058}}\textbf{$\pm$}0.013 & 0.091\textbf{$\pm$}0.013 & \textcolor{c2}{\textbf{0.084}}\textbf{$\pm$}\textcolor{c3}{\textbf{0.002}} & \textcolor{c2}{\textbf{0.057}}\textbf{$\pm$}\textcolor{c2}{\textbf{0.001}} & \textcolor{c2}{\textbf{0.132}}\textbf{$\pm$}0.035 & 0.017\textbf{$\pm$}0.003 & 0.093\textbf{$\pm$}0.030 \\
		AE7 & \textcolor{c3}{\textbf{0.028}}\textbf{$\pm$}\textcolor{c3}{\textbf{0.000}} & 0.071\textbf{$\pm$}\textcolor{c2}{\textbf{0.002}} & \textcolor{c2}{\textbf{0.021}}\textbf{$\pm$}\textcolor{c3}{\textbf{0.000}} & 0.066\textbf{$\pm$}0.010 & 0.069\textbf{$\pm$}0.013 & 0.089\textbf{$\pm$}0.004 & 0.059\textbf{$\pm$}0.002 & 0.195\textbf{$\pm$}0.036 & \textcolor{c2}{\textbf{0.016}}\textbf{$\pm$}\textcolor{c3}{\textbf{0.001}} & \textcolor{c3}{\textbf{0.083}}\textbf{$\pm$}\textcolor{c3}{\textbf{0.003}} \\
		AE8 & \textcolor{c3}{\textbf{0.028}}\textbf{$\pm$}\textcolor{c3}{\textbf{0.000}} & 0.067\textbf{$\pm$}\textcolor{c2}{\textbf{0.002}} & \textcolor{c2}{\textbf{0.021}}\textbf{$\pm$}\textcolor{c3}{\textbf{0.000}} & 0.067\textbf{$\pm$}\textcolor{c2}{\textbf{0.009}} & 0.086\textbf{$\pm$}0.017 & \textcolor{c3}{\textbf{0.082}}\textbf{$\pm$}\textcolor{c2}{\textbf{0.003}} & \textcolor{c2}{\textbf{0.057}}\textbf{$\pm$}0.002 & 0.143\textbf{$\pm$}0.040 & 0.017\textbf{$\pm$}0.004 & 0.100\textbf{$\pm$}0.044 \\
		AE9 & 0.045$\pm$0.020 & 0.090$\pm$0.007 & 0.054$\pm$0.009 & 0.085$\pm$0.017 & 0.090$\pm$0.020 &  0.134$\pm$0.035 & 0.150$\pm$0.009 & 0.172$\pm$0.022 & 0.023$\pm$0.006 & 0.095$\pm$0.014 \\
		Ours & \textcolor{c3}{\textbf{0.028}}\textbf{$\pm$}\textcolor{c3}{\textbf{0.000}} & \textcolor{c2}{\textbf{0.066}}\textbf{$\pm$}\textcolor{c3}{\textbf{0.001}} & \textcolor{c3}{\textbf{0.020}}\textbf{$\pm$}\textcolor{c3}{\textbf{0.000}} & \textcolor{c3}{\textbf{0.051}}\textbf{$\pm$}\textcolor{c3}{\textbf{0.006}} & \textcolor{c3}{\textbf{0.061}}\textbf{$\pm$}\textcolor{c2}{\textbf{0.002}} & 0.086\textbf{$\pm$}0.004 & \textcolor{c3}{\textbf{0.056}}\textbf{$\pm$}\textcolor{c2}{\textbf{0.001}} & \textcolor{c3}{\textbf{0.121}}\textbf{$\pm$}0.006 & \textcolor{c3}{\textbf{0.015}}\textbf{$\pm$}\textcolor{c3}{\textbf{0.001}} & \textcolor{c3}{\textbf{0.083}}\textbf{$\pm$}\textcolor{c2}{\textbf{0.005}} \\
		
		\hline
	\end{tabular}
	\label{tab:ablation}
\end{table*}
\subsection{Ablation Study}
\label{exp_ablation}
The performance of our proposed 3D-CSCNet depends on the network architecture, endmember initialization, and the two-stage training procedure. Especially, our proposed 3D-CSCB effectively estimates a 3D sparse feature representation from the HSI, which is then used to estimate the preliminary abundance matrix. Moreover, we propose PSVM algorithm to extract an initial estimate of the endmembers, which are then utilized to initialize the decoder weights of our 3D-CSCNet. Further, we propose a two-stage training procedure, where we train only the encoder parameters while keeping the well-initialized decoder layer fixed in training Stage I. Then, in training Stage II, we train the parameters of both the encoder and decoder layers, ensuring fine-tuning of both the abundance and endmembers. In this section, we conduct an ablation study to verify the effectiveness of our proposed method. First, we verify the effectiveness of our PSVM algorithm with other endmember estimation methods, and the results are presented in Table \ref{tab:algorithms}. Then, we study the effects of the ablation experiments on the HSU performance, and these results are presented in Table \ref{tab:ablation}.

\noindent \textbf{Effectiveness of PSVM algorithm: }To verify the effectiveness of our PSVM algorithm, we compare it's performance in estimating endmembers with three other algorithms: VCA \cite{vca}, SVM \cite{simplex1}, and $\text{PSVM}_\text{ND}$. Here, $\text{PSVM}_\text{ND}$, a variant of our PSVM algorithm, does not have the conditional denoising step. As the results in Table \ref{tab:algorithms} show, VCA has significant variations across different runs. PSVM has improved performance, specifically in the presence of noise.

\noindent \textbf{Effectiveness of 3D-CSC model: }Our novel 3D-CSCB estimates 3D-sparse feature from which the abundance matrix is estimated. To validate the effectiveness of the 3D-CSC, we conduct an ablation experiment, denoted as \textbf{AE1}. Here, instead of estimating 3D sparse feature, we estimate 2D sparse feature. As the results in Table \ref{tab:ablation} show, estimating 3D sparse feature significantly improves the HSU performance.

\noindent \textbf{Effectiveness of two-stage training procedure: }To verify the effectiveness of our proposed two-stage training procedure, we conduct an ablation experiment (denoted as \textbf{AE2}), where we train both the encoder and decoder in one stage training. As the results show, training 3D-CSCNet in two stages has better overall performance.

\noindent \textbf{Effects of sharing convolution layers in 3D-CSCB: }We use separate convolution layers in different IMs of 3D-CSCB. This design is motivated from \cite{maximal}, which suggests that employing separate learnable layers at different iterations improves the sparse feature estimation. To validate this design choice, we conduct an ablation experiment (denoted as \textbf{AE3}), where we share the convolution layers across iterations of 3D-CSCB. As the results show, sharing convolution layers across iterations degrades the performance.

\noindent \textbf{Effectiveness of constraining the threshold values in 3D-CSCB: }Motivated by \cite{sinet}, we constrain the threshold values $\theta^{(k)}$ to be non-negative and monotonically decreasing with the iteration index $k$, since the noise level in $z$ decreases over iterations, as presented in Equation \ref{eq_constrain}. To validate this design choice, we conduct an ablation experiment (denoted as \textbf{AE4}), where we do not constrain the threshold values in 3D-CSCB. As the results show, constraining the threshold values significantly improves the performance.

\noindent \textbf{Effects of changing number of IMs in 3D-CSCB: }We use six IMs in our 3D-CSCB. To analyze the effects of changing number of IMs in 3D-CSCB, we conduct two ablation experiments. \textbf{AE5}: we use five IMs in 3D-CSCB, and \textbf{AE6}: we use seven IMs in 3D-CSCB. As the results show, employing six IMs in 3D-CSCB gives the best overall performance.

\noindent \textbf{Effects of changing number of filters in 3D-CSCB: }We use $48$ number of filters in the convolution layers of our 3D-CSCB. To analyze the effects of changing number of filters, we conduct two ablation experiments. \textbf{AE7}: we use $32$ number of filters, and \textbf{AE8}: we use $64$ number of filters. As the results show, employing $48$ number of filters in the convolution layers of 3D-CSCB gives the best overall performance.

\noindent \textbf{Effectiveness of endmember initialization: }We employ our proposed PSVM algorithm to extract the initial endmembers, which are used to initialize the decoder weights of 3D-CSCNet. To verify the effectiveness of this endmember initialization, we conduct an ablation experiment (denoted as \textbf{AE9}), where we employ VCA to initialize the decoder weights. As the results show, initializing with PSVM significantly improves the performance.

The results in Table \ref{tab:ablation} show that 3D-CSCNet has the best overall performance compared to the ablation experiments, which demonstrates the effectiveness of our proposed method.

\section{Conclusion}
\label{conclusion}
In this work, we introduced 3D-CSCNet, a novel algorithm-unrolling-based network designed for hyperspectral unmixing. By embedding a 3D convolutional sparse coding model into an autoencoder framework, 3D-CSCNet effectively combines the advantages of optimization-based CSC with deep learning. The proposed 3D-CSC block (3D-CSCB) unrolls the 3D CSC optimization process to estimate a spatial–spectral sparse representation, enabling accurate abundance estimation while fully exploiting the 3D structure of hyperspectral data. Moreover, we developed the PSVM algorithm, which provides improved performance in estimating endmembers. These endmembers are used to initialize the decoder weights, and coupled with our two-stage training strategy, allow the network to first learn the abundances and subsequently fine-tune both abundances and endmembers in an unsupervised manner. Extensive experiments on three real datasets and one simulated dataset across three SNR levels demonstrate that 3D-CSCNet consistently surpasses SOTA unmixing methods in both abundance estimation and endmember extraction. Ablation studies further validate the contributions of the 3D-CSCB, PSVM initialization, and the two-stage training process. Future work could explore more unrolling-based AE networks for the HSU task.

\bibliographystyle{IEEEtran}
\bibliography{main}

\end{document}